\documentclass[journal]{IEEEtran}

\usepackage{cite}

\ifCLASSINFOpdf
  \usepackage[pdftex]{graphicx}
  \DeclareGraphicsExtensions{.eps,.pdf,.jpeg,.png}
\else
  \usepackage[dvips]{graphicx}
  \DeclareGraphicsExtensions{.eps,.pdf}
\fi
\graphicspath{{./}}

\usepackage{amsmath}
\usepackage{bm}
\usepackage{amssymb} 
\usepackage{amsthm}
\newtheorem{thm}{Theorem}

\newtheorem*{defin}{Problem Definition}

\DeclareMathOperator*{\argmin}{argmin}

\usepackage{algorithmic}
\usepackage{algorithm}

\newcommand{\PARAMS}{\textbf{Parameter:}}

\usepackage{array}
\usepackage{multirow}
\usepackage{threeparttable}
\usepackage{booktabs}

\ifCLASSOPTIONcompsoc
  \usepackage[caption=false,font=normalsize,labelfont=sf,textfont=sf]{subfig}
\else
  \usepackage[caption=false,font=footnotesize]{subfig}
\fi

\usepackage{tikz}
\usetikzlibrary{fit}
\usetikzlibrary{backgrounds}
\usetikzlibrary{fadings}
\usetikzlibrary{positioning}

\begin{document}



\newcommand{\citep}{\cite}
\newcommand{\citet}[1]{\citeauthor{#1} \shortcite{#1}}
\newcommand{\citealp}[1]{\citeauthor{#1} \citeyear{#1}}

\def\entitle{Sub-Architecture Ensemble Pruning in Neural Architecture Search}
\def\propfull{\entitle{}}
\def\propabbr{\emph{SAEP}}
\def\randfull{Pruning by Random Selection}
\def\valufull{Pruning by Accuracy Performance}
\def\infofull{Pruning by Information Entropy}
\def\randabbr{\emph{PRS}}
\def\valuabbr{\emph{PAP}}
\def\infoabbr{\emph{PIE}}
\def\adanet{AdaNet}
\def\variant{\emph{.W}}
\def\randvari{\randabbr\variant}
\def\valuvari{\valuabbr\variant}
\def\infovari{\infoabbr\variant}
\def\adanvari{\adanet\variant}
\def\propvari{\propabbr\variant}

\def\nasfull{neural architecture search}
\def\nasabbr{NAS}
\def\complexity{Rademacher complexity}
\def\etal{\emph{et al.}}
\def\eg{e.g.,}
\def\ie{i.e.,}
\def\batchnorm{\mathrm{BN}}
\def\convolution{\mathrm{Conv}}
\def\relu{\mathrm{ReLU}}
\def\convexhull{\mathrm{ConvexHull}}
\def\Error{\mathrm{Err}}
\def\ii{\mathrm{I}}
\def\hh{\mathrm{H}}
\def\mi{\mathrm{MI}}
\def\vi{\mathrm{VI}}
\def\dist{\mathrm{TDAC}}
\def\divs{\mathrm{TDAS}}
\def\disagreement{\mathrm{dis}}


\title{\entitle{}}
%
%
%

%
%
\author{Yijun~Bian,
    Qingquan~Song,
    Mengnan~Du,
    Jun~Yao,
    Huanhuan Chen,~\IEEEmembership{Senior~Member,~IEEE,}
    and~Xia~Hu
\thanks{Y. Bian and H. Chen are with the School of Computer Science and Technology, University of Science and Technology of China, Hefei 230027, China. E-mails: yjbian@mail.ustc.edu.cn; hchen@ustc.edu.cn}%
\thanks{Q. Song, M. Du, and X. Hu are with the Department of Computer Science and Engineering, Texas A\&M University, College Station, TX, 77840, United States. E-mails: song\_3134@tamu.edu; dumengnan@tamu.edu; hu@cse.tamu.edu}%
\thanks{J. Yao is with the Data Science and Analytics Department, WeBank, Shenzhen, 518000, China. E-mail: junyao@webank.com}
\thanks{Manuscript received December 05, 2019; revised May 17, 2020; revised February 12, 2021; accepted May 18, 2021. This research is supported in part by the National Key Research and Development Program of China under Grant No. 2016YFB1000905, the National Natural Science Foundation of China under Grant No. 91746209, and the Fundamental Research Funds for the Central Universities. Corresponding author: Huanhuan Chen.}%
}

\markboth{IEEE Transactions on Neural Networks and Learning Systems}%
{Bian \MakeLowercase{\textit{et al.}}: \entitle{}}
%



\maketitle
\begin{abstract}
Neural architecture search (NAS) is gaining more and more attention
in recent years due to its flexibility and remarkable capability 
to reduce the burden of neural network design. To achieve better
performance, however, the searching process usually costs massive
computations that might not be affordable for researchers and
practitioners. While recent attempts have employed ensemble learning
methods to mitigate the enormous computational cost, however, 
they neglect a key property of ensemble methods, namely diversity, 
which leads to collecting more similar sub-architectures with potential
redundancy in the final design. To tackle this problem, we propose a
pruning method for NAS ensembles called ``\emph{\propfull{}
(}\propabbr{}\emph{)}.'' It targets to leverage diversity and to achieve
sub-ensemble architectures at a smaller size with comparable
performance to ensemble architectures that are not pruned. Three
possible solutions are proposed to decide which sub-architectures
to prune during the searching process. Experimental results
exhibit the effectiveness of the proposed method by largely
reducing the number of sub-architectures without degrading the
performance.
\end{abstract}

%
\IEEEpeerreviewmaketitle

\section{Introduction}
\label{intro}

\IEEEPARstart{D}{esigning} neural network architectures usually
requires manual, laborious architectural engineering, extensive
expertise, and high costs. Neural architecture search
(NAS), which aims to mitigate these challenges, is attracting increasing 
attention recently \citep{zoph2018learning,elsken2019neural,wistuba2019survey}.
However, NAS methods usually require a huge computational effort to achieve an
architecture with the expected performance, which is too expensive for
many infrastructures and too costly for researchers \citep{zoph2017neural}. 
Recent work
\citep{cortes2017adanet,huang2018learning,macko2019improving}
proposes to employ ensemble methods to mitigate this shortcoming 
by combining weak sub-architectures trained with lower computational cost into powerful neural architectures. 
\adanet{}, as a prominent example of them, contributes to present a theoretical analysis of the problem of learning both the network architecture and its parameters simultaneously, and proposes the first generalization bounds for the problem of structural learning of neural networks \citep{cortes2014deep,cortes2017adanet}. 

However, all of them 
overlook a crucial principle in ensemble methods (\ie{} model
diversity) in the search for new sub-architectures, which is
usually beneficial for creating better model ensembles
\citep{chen09,chen10,chen2007evolutionary}. Besides, lots of
ensemble pruning methods exploit the diversity property to
obtain sub-ensembles with a smaller size than the original ensembles
\citep{bian2019ensemble,chen06probabilistic}. It has been proved
that a few diverse individual learners could even construct a
more powerful ensemble learner than the unpruned ensembles
\citep{chen2009predictive,chen2008diversity}. This motivates us to
investigate the NAS ensemble pruning problem, where different sub-ensemble architectures are aligned to a smaller but effective ensemble model. 
Moreover, it is quite challenging to describe the characteristics of
diversity for different sub-architectures and decide which one of them should be 
pruned or kept in the ensemble architecture. First, there are plenty
of definitions or measurements for diversity in the ensemble
learning community \citep{chen2008diversity}. Unlike the model
accuracy, however, there is no well-accepted formal definition of
diversity \citep{bian2019does}. Second, diversity among individual
learners usually decreases as those individual learners approach higher levels of accuracy \citep{lu2010ensemble}. Combining some diverse individual learners with some relatively weak ones is
usually better than combining accurate ones only since diversity is
more important than pure accuracy. Third, selecting the best
combination of sub-architectures from an ensemble architecture is
NP-complete hard with exponential computational complexity
\citep{li2012diversity,martinez2007using}. Thus, how to manage the
trade-off between accuracy and diversity properly, and how to select the
best subset of ensemble architectures, is a significant problem in the
NAS ensemble pruning problems.

Motivated by the characteristic of diversity in ensemble learning, we strive for diverse sub-ensemble architectures at a smaller size, meanwhile, maintaining comparable accuracy performance to the original ensemble architecture without pruning. 
The idea is to prune the ensemble architecture on-the-fly
based on various criteria and keep more valuable sub-architectures
in the searching process. Our NAS ensemble pruning method is named
as ``\emph{\propfull{} (}\propabbr{}\emph{)},'' motivated by AdaNet
\citep{cortes2017adanet} and ensemble pruning methods, with three proposed criteria to decide
which sub-architectures to prune. 
Note that \propabbr{} also has some differences from typical ensemble pruning problems, since most pruning methods are usually handled on the original ensemble that has been trained, but pruning in \propabbr{} is done in the process of searching rather than after the searching. 
Moreover, \propabbr{} might
lead to distinct deeper architectures than the
original one if the degree of diversity is insufficient, which could
be a bonus due to pruning. Our contribution in this paper is threefold:
\begin{itemize}
\item We propose a NAS ensemble pruning method to search sub-ensemble architectures at a smaller size that benefits from an essential characteristic, \ie{} diversity in ensemble learning. It could achieve comparable accuracy performance to ensemble architectures that are not pruned.
\item Moreover, our proposed method would lead to distinct deeper architectures than the original ensemble architecture that is not pruned if the diversity is insufficient.
\item Experimental results exhibit the effectiveness of the proposed method in largely reducing the number of sub-architectures in ensemble architectures and increasing the diversity while maintaining the final performance.
\end{itemize}

\section{Problem Statement}

\textbf{Notations}: %
In this paper, we denote tensors with bold italic lowercase letters (\eg{} $\bm{x}$), vectors with bold lowercase letters (\eg{} $\mathbf{x}$), and scalars with italic lowercase letters (\eg{} $x$).
We use $\mathbf{x}^\mathsf{T}$ to represent the transpose of a vector.
Data/hypothesis spaces are denoted by bold script uppercase letters (\eg{} $\mathcal{X}$).
We use $\mathbb{R,P,E}$, and $\mathbb{I}$ to denote the real space, the probability measure, the expectation of a random variable, and the indicator function, respectively.

We summarize the notations and their definitions in
Table~\ref{tab:notation}. We follow the notations and the definition
of the search space in \adanet{} to formulate the problem and introduce
the proposed method, as it is one of the most popular ensemble
search methods in the NAS literature. It is worth mentioning that
the proposed pruning criteria could also be generalized to other
ensemble methods, which could be interesting for future research.

Let $f$ be a neural network  with $l$ layers searched via \adanet{}
\citep{cortes2014deep,cortes2017adanet}, where each layer would
be connected to the previous layers. The output for each
$\bm{x}\in\mathcal{X}$ would connect to all intermediate units,
\ie{}
\begin{equation}
\small
    f(\bm{x})= \sum_{1\leqslant k\leqslant l}
    \mathbf{w}_k\cdot\mathbf{h}_k(\bm{x}) \,,\label{eq:1}
\end{equation}
where $\sum_{k=1}^l\|\mathbf{w}_k\|_1=1$ and $\mathbf{h}_k=[h_{k,1},...,h_{k,n_k}]^\mathsf{T}$.
$h_{k,j}$ is the function of a unit in the $k^\text{th}$ layer, \ie{} %
\begin{equation}
\small
    h_{k,j}(\bm{x})= \sum_{0\leqslant s\leqslant k-1}
    \mathbf{u}_s\cdot\phi_s\big( \mathbf{h}_s(\bm{x}) \big)
    \,,\; k\in[l]
    \,, \label{eq:2}
\end{equation}
where $h_0(\bm{x})=\bm{x}$ is the $0^\text{th}$ layer denoted by the
input. Note that $\phi_s(\mathbf{h}_s)$ denotes that
$\phi_s(\mathbf{h}_s)=\big(\phi_s(h_{s,1}),...,\phi_s(h_{s,n_s})\big)$
where the $\phi_s$ is assumed to be 1-Lipschitz activation
functions, such as the ReLU\footnote{The Rectified Linear function
(ReLU function)
\citep{jarrett2009best,nair2010rectified,glorot2011deep} is defined
as $g(z)= \max\{0,z\}$.} or sigmoid\footnote{The sigmoid function
\citep{goodfellow2016deep1} is defined as $\sigma(z)=
\frac{1}{1+e^{-z}}$.} function \citep{cortes2017adanet}.
If $\mathbf{u}_s=0$ for $s<k-1$ and $\mathbf{w}_k=0$ for $k<l$, this
architecture of $f$ will coincide with the standard multi-layer
feed-forward ones \citep{cortes2017adanet}.

\begin{table}[t]
\centering\small
\caption{The used symbols and definitions in this paper.}
\label{tab:notation}
\scalebox{0.83}{%
\begin{tabular}{ll}
    \toprule
    Notation & Definition \\
    \midrule
    $[n]$ & the representation of $\{1,...,n\}$ for clarity \\
    $\bm{x}\in\mathcal{X}$ & the input of neural networks \\
    $f(\cdot)\in\mathcal{F}$ & the function of a neural network with $l$ layers \\
    $n_s$ & the number of units in the $s^\text{th}$ layer \\
    $h_{k,j}(\cdot)$ & the function of a unit in the $k^\text{th}$ layer $(k\in[l])$ \\
    $\mathbf{u}_s\in\mathbb{R}^{n_s}$ & the weight of the $s^\text{th}$ layer for the units of the $k^\text{th}$ layer \\
    $\mathbf{h}_k(\cdot)$ & the function vector of units in the $k^\text{th}$ layer \\
    $\mathbf{w}_k\in\mathbb{R}^{n_k}$ & the weight of the $k^\text{th}$ layer for $f(\cdot)$ \\
    $\|\mathbf{w}_k\|_p$ & the $l_p$-norm of $\mathbf{w}_k$ where $p\geqslant 1$ \\
    $T\geqslant 1$ & the number of iterations in the neural architecture search- \\ & ing process \\
    $\Gamma$ & a specific complexity constraint based on the Rademacher \\ & complexity \\
    \bottomrule
\end{tabular}
}
\end{table}

To investigate the search space $\mathcal{F}$, $\mathcal{H}_k$ is used to denote the family of the function in the $k^\text{th}$ layer.
Let $\widetilde{\mathcal{H}}_k\overset{\text{def}}{=} \mathcal{H}_k\cup(-\mathcal{H}_k)$ denote the union of $\mathcal{H}_k$ and its reflection, and let $\mathcal{H}\overset{\text{def}}{=} \cup_{k=1}^l\widetilde{\mathcal{H}}_k$ denote the union of the families $\widetilde{\mathcal{H}}_k$.
Then $\mathcal{F}$ coincides with the convex hull of $\mathcal{H}$, which means that generalization bounds for ensemble methods could be utilized to analyze learning with $\mathcal{F}$ \citep{cortes2017adanet}.
Therefore, Cortes~\etal{}~\citep{cortes2014deep,cortes2017adanet} 
attempted to propose learning guarantees based on a \complexity{}
analysis \citep{koltchinskii2002empirical}  to guide their design of
algorithms.

While AdaNet attempts to train multiple weak sub-architectures with lower computational costs to comprise powerful neural architectures inspired by ensemble methods \citep{cortes2017adanet},
the crucial characteristic of diversity brings opportunities to achieve sub-ensemble architectures at a smaller size with diverse sub-architectures, yet still with the comparable performance to an original ensemble architecture generated by AdaNet.
Based on the above notions, we formally define the NAS ensemble pruning problem.

\begin{defin}[NAS Ensemble Pruning]
\label{def:nas_ens_pru}
Given an ensemble architecture $f(\bm{x})=\sum_{1\leqslant  k\leqslant  l}\mathbf{w}_k\cdot\mathbf{h}_k(\bm{x})\in\mathcal{F}$ searched by ensemble NAS methods such as AdaNet, and a training set $S=\{(\bm{x}_1,y_1),...,(\bm{x}_m,y_m)\}$ with the size of $m$, 
assuming that all training instances are drawn i.i.d. (independently and identically distributed) from a distribution $\mathcal{D}$ over 
$\mathcal{X}\times\{c_1,...,c_{n_c}\}$ with $n_c$ as the number of labels,
the goal is to prune the ensemble architecture $f$ and search for a sub-ensemble architecture of a smaller size, while maintaining comparable performance to the original ensemble architecture $f$.
\end{defin}

\section{\propfull{} (\propabbr{})}
\label{methods}

In this section, we elaborate on the proposed NAS ensemble pruning
method to obtain smaller yet effective neural ensemble
architectures. Before pruning the less valuable sub-architectures,
we need to generate sub-architectures first. We take advantage of
AdaNet \citep{cortes2017adanet} here due to its popularity and
superiority in ensemble NAS research, and utilize its objective
function to generate candidate sub-architectures in the searching
process. The objective function for generating new candidates in AdaNet
is defined as
\begin{equation}
\small
    \mathcal{L}_g(\mathbf{w})= \hat{R}_{S,\rho}(f)+ \Gamma
    \,,\label{eq:6}
\end{equation}
where $\hat{R}_{S,\rho}(f)$ denotes the empirical margin error of function $f$ on the training set $S$, and $\Gamma$ denotes a specific complexity constraint.

As the learning guarantee in \citep{cortes2017adanet} applies to binary classification, we introduce an auxiliary function $g(\bm{x},y,f)$ in Eq.~(\ref{eq:3}) to extend the objective to multi-class classification problems consistent with our problem statement, \ie{}
\begin{equation}
\small
    g(\bm{x},y,f)= 2\mathbb{I}\big(
        f(\bm{x})=y
    \big)-1 \,.\label{eq:3}
\end{equation}
In this case, the empirical margin error $\hat{R}_{S,\rho}(f)$ would be
\begin{equation}
\small
    \hat{R}_{S,\rho}(f)= \frac{1}{m}\sum_{1\leqslant i\leqslant m}\mathbb{I}\big( g(\bm{x}_i,y_i,f)\leqslant\rho \big)
    \,.\label{eq:4}
\end{equation}

\begin{figure}
\centering
    \subfloat[]{\label{fig:incre2:adanet}
        \includegraphics[scale=0.27]{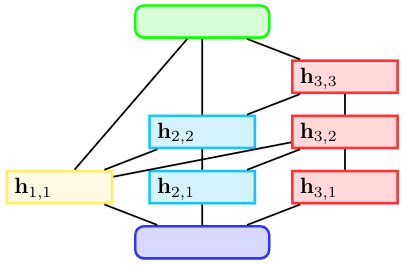}}
    \hspace{0.05in}
    \subfloat[]{\label{fig:incre2:respru}
        \includegraphics[scale=0.27]{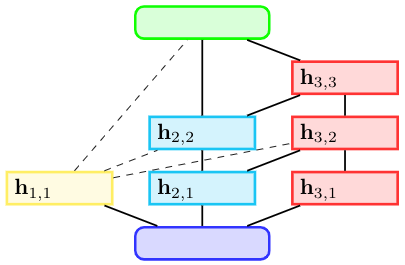}}
    \hspace{0.05in}
\caption{%
This figure is used to illustrate the difference between \propabbr{} and \adanet{} during the incremental construction of neural architectures. 
Layers in blue and green indicate the input and output layers, respectively.
Units in yellow, cyan, and red are added at the first, second, and third iteration, respectively.
(a) AdaNet~\protect\citep{cortes2017adanet}:
A line between two blocks of units indicates that these blocks are fully-connected.
(b) \propabbr{}: Only some valuable blocks are kept (those that will be pruned are denoted by black dashed lines), which is the key difference from AdaNet.
The criteria used to decide which sub-architectures will be pruned have three proposed solutions in our \propabbr{}, \ie{} \randabbr{}, \valuabbr{}, and \infoabbr{}.
}\label{fig:incre}
\end{figure}

Guided by Eq.~\eqref{eq:6}, AdaNet only generates new candidates by minimizing the empirical error and architecture complexity, while overlooking the diversity and differences among different sub-architectures. To achieve smaller yet effective ensembles via taking the diversity property into account, we need first to measure the diversity of different sub-architectures so that a corresponding objective function could be derived to guide us for the selection of more valuable sub-architectures during the searching process.

Specifically, we propose three different ways to enhance the diversity of different sub-architectures. Except for the first solution, the latter two provide specific objective quantification where diversity is involved as guidance among different sub-architectures for NAS. Besides, the diversity of sub-ensemble architectures generated by them could be quantified to verify whether these ways work or not.

Our final NAS ensemble pruning method, named as ``\emph{\propfull{} (}\propabbr\emph{)},'' is shown in Algorithm~\ref{alg:prop}.
The key difference between \propabbr{} and \adanet{} is that \propabbr{}  prunes the less valuable sub-architectures based on certain criteria during the searching process (lines \ref{alg:line:10}--\ref{alg:line:13} in Algorithm~\ref{alg:prop}), instead of keeping all of them, as shown in Figure~\ref{fig:incre}.
At the $t^\text{th}$ iteration $(t\in[T])$ in Algorithm~\ref{alg:prop}, let $f^{(t-1)}=\sum_{1\leqslant k\leqslant l}\mathbf{w}_k\cdot\mathbf{h}_k$ denote the neural network constructed before the start of the $t^\text{th}$ iteration, with the depth $l^{(t-1)}$ of $f$.
The first target at the $t^\text{th}$ iteration is to generate new candidates (lines \ref{alg:line:3}--\ref{alg:line:4}) and select the better one to be added in the model of $f^{(t-1)}$ (lines \ref{alg:line:4}--\ref{alg:line:9}) since we expect the searching process is progressive.
The second target at the $t^\text{th}$ iteration is to prune the less valuable sub-architectures for $f^{(t)}$ and keep beneficial ones to construct the final architecture (lines \ref{alg:line:10}--\ref{alg:line:13}).

\begin{algorithm}[t!]
\small\caption{%
\propfull{} (\propabbr{})}\label{alg:prop}
\begin{algorithmic}[1]
\REQUIRE Dataset $S=(\bm{x}_i,y_i)_{i=1}^m$ \\
\PARAMS{} Number of iteration $T$
\ENSURE Final function $f^{(T)}$
    \STATE Initialize $f^{(0)}=\bm{0} \,,$ and $l^{(0)}=1 \,.$
    \FOR{$t=1$ \textbf{to} $T$}
        \STATE $\mathbf{w}',\mathbf{h}'=\argmin_{\mathbf{w},\mathbf{h}}\mathcal{L}_g( f^{(t-1)}+\mathbf{w}\cdot\mathbf{h} )$ s.t. $\mathbf{h}\in\mathcal{H}_{l^{(t-1)}} .$
        \label{alg:line:3}
        \STATE $\mathbf{w}'',\mathbf{h}''=\argmin_{\mathbf{w},\mathbf{h}}\mathcal{L}_g( f^{(t-1)}+\mathbf{w}\cdot\mathbf{h} )$ s.t. $\mathbf{h}\in\mathcal{H}_{l^{(t-1)}+1} .$
        \label{alg:line:4}
        \IF{$ \mathcal{L}_g(f^{(t-1)}+\mathbf{w}'\cdot\mathbf{h}') \leqslant \mathcal{L}_g(f^{(t-1)}+\mathbf{w}''\cdot\mathbf{h}'') $}
        \label{alg:line:5}
            \STATE $f^{(t)}= f^{(t-1)}+ \mathbf{w}'\cdot\mathbf{h}' .$
        \ELSE
            \STATE $f^{(t)}= f^{(t-1)}+ \mathbf{w}''\cdot\mathbf{h}'' .$
        \ENDIF
        \label{alg:line:9}
        \STATE \emph{Choose $\mathbf{w}_p$ based on one certain criterion, \ie{} picking randomly in \randabbr{}, $\mathcal{L}_d(\mathbf{w})$ of Eq.~(\ref{eq:7b}) in \valuabbr{}, or $\mathcal{L}_e(\mathbf{w}_i)$ of Eq.~(\ref{eq:11}) in \infoabbr{}. }
        \label{alg:line:10}
        \STATE \emph{Set $\mathbf{w}_p$ to be zero. }
        \label{alg:line:13}
    \ENDFOR
\end{algorithmic}
\end{algorithm}

To evaluate the most valuable sub-architectures, we propose three solutions to tackle this problem.
Now we introduce them to decide which sub-architectures are less valuable to be pruned.

\subsection{\randfull{} (\randabbr{})}
\label{prop:a}

The first solution, named as ``\emph{\randfull{} (}\randabbr{}\emph{)},'' is to randomly prune some of the sub-architectures in the searching process, with one difference from other solutions.
In \randabbr{}, we firstly decide randomly whether or not to pick one of the sub-architectures to be pruned;
if we indeed decide to prune one of them, the objective to decide which sub-architectures to prune is random as well, instead of the specific objective in the next two solutions.

However, there is no specific objective for \randabbr{} to follow in the pruning process.
That might lead to a situation where some valuable sub-architectures are pruned as well.
Therefore, we need to find more explicit objectives to guide our pruning.

\subsection{\valufull{} (\valuabbr{})}
\label{prop:b}

To measure different sub-architectures better, we propose the second pruning solution based on their accuracy performance.
This method is named as ``\emph{\valufull{} (}\valuabbr{}\emph{)}.''
To choose the valuable sub-architectures from those individual sub-architectures in the original model, this second optional objective function for this target is defined as
\begin{equation}
\small
    \mathcal{L}_d(\mathbf{w})= \frac{1}{m}\sum_{1\leqslant i\leqslant m}[
        g(\bm{x}_i,y_i,f)-g(\bm{x}_i,y_i,f-\mathbf{w}\cdot\mathbf{h})
    ] \,,\label{eq:7b}
\end{equation}
where $\mathbf{h}$ is the sub-architecture corresponding to the
weight $\mathbf{w}$. The target is to pick up the $\mathbf{w}$ and
$\mathbf{h}$ by minimizing Eq.~(\ref{eq:7b}), and prune them if
their loss is less than zero. The reason why we do this is that the
generalization error of gathering all sub-architectures is defined
as
\begin{equation}
\small
    R(f)= \mathbb{E}_{(\bm{x},y)\sim\mathcal{D}}\big[\mathbb{I}\big( g(\bm{x},y,f) \leqslant 0 \big)\big]
    \,;\label{eq:5}
\end{equation}
if the $j^\text{th}$ sub-architecture is excluded from the final architecture, the generalization error of the pruned sub-ensemble architecture will become
\begin{equation}
\small
    R(\bar{f}_j)= \mathbb{E}_{(\bm{x},y)\sim\mathcal{D}}[
        \mathbb{I}(
            g(\bm{x},y,f-\mathbf{w}_j\cdot\mathbf{h}_j) \leqslant 0
    )]\,.\label{eq:8}
\end{equation}
Then, if we expect the pruned architecture works better than the original one, we need to make sure that $R(f)-R(\bar{f}_j)\geqslant 0$, \ie{}
\begin{equation}
\small
    \mathbb{E}_{(\bm{x},y)\sim\mathcal{D}}\big[
        g(\bm{x},y,f)-g(\bm{x},y,f-\mathbf{w}_j\cdot\mathbf{h}_j)
    \big]\leqslant 0 \,.\label{eq:9}
\end{equation}
Therefore, if the $j^\text{th}$ sub-architecture meeting Eq.~\eqref{eq:9} is excluded from the final architecture, the performance will not be weakened and could be even better than the original one.
The hidden meaning behind Eq.~\eqref{eq:9} is that the final architecture makes mistakes; however, the pruned architecture that excludes the $j^\text{th}$ sub-architecture will work correctly.
These sub-architectures that make too serious mistakes to affect the final architecture negatively would be expected to be pruned, leading to our loss function Eq.~\eqref{eq:7b}.
In this case, we could \emph{improve the performance of the final architecture without breaking the learning guarantee}.

However, this objective in Eq.~(\ref{eq:7b}) only considers the accuracy performance of different sub-architectures and misses out on the crucial characteristic of diversity in ensemble methods.
Therefore, we need to find an objective to reflect accuracy and diversity both.

\subsection{\infofull{} (\infoabbr{})}
\label{prop:c}

To consider accuracy and diversity simultaneously, we propose another strategy, named ``\emph{\infofull{} (}\infoabbr{}\emph{)}.''  The objective is based on information entropy.
For any sub-architecture $\mathbf{w}_j$ in the ensemble architecture, $\bm{w}_j=\mathbf{w}_j\cdot[\mathbf{h}_j(\bm{x}_1),...,\mathbf{h}_j(\bm{x}_m)]^\mathsf{T}$ represents its classification results on the dataset $S$.
$\mathbf{y}=[y_1,...,y_m]^\mathsf{T}$ is the class label vector.
Notice that $H(\cdot)$ and $H(\cdot,\cdot)$ are the entropy function and the joint entropy function, respectively, \ie{}
\begin{small}
\begin{align}
    H(\bm{w}_i) =& -\sum_{w\in \bm{w}_i} p(w)\log p(w) \,,\\
    H(\bm{w}_i,\mathbf{y})=& -\sum_{w\in \bm{w}_i}\sum_{y\in \mathbf{y}} p(w,y)\log p(w,y) \,.
\end{align}%
\end{small}%
To exhibit the relevance between this sub-architecture and the class label vector, the normalized mutual information \citep{zadeh2017scalable},
\begin{small}
\begin{align}
    \mi(\bm{w}_i,\mathbf{y}) =& \frac{ \ii(\bm{w}_i;\mathbf{y}) }{ \sqrt{ \hh(\bm{w}_i)\hh(\mathbf{y}) } } \nonumber\\
    =& \frac{ \sum_{w\in\bm{w}_i, y\in\mathbf{y}} p(w,y)\log\frac{p(w,y)}{p(w)p(y)} }{\sqrt{ \sum_{w\in\bm{w}_i}p(w)\log p(w) \sum_{y\in\mathbf{y}}p(y)\log p(y) }} \,,
\end{align}%
\end{small}%
is used to imply its accuracy.
Note that
\begin{small}
\begin{align}
    \ii(\bm{w}_i;\mathbf{y})=& \hh(\bm{w}_i)-\hh(\bm{w}_i|\mathbf{y}) \nonumber\\
    =& \sum_{w\in \bm{w}_i,y\in\mathbf{y}} p(w,y)\log\frac{p(w,y)}{p(w)p(y)} \,,
\end{align}%
\end{small}%
is the mutual information \citep{cover2012elements}.
To reveal the redundancy between two sub-architectures ($\mathbf{w}_i$ and $\mathbf{w}_j$) in the ensemble architecture, the normalized variation of information \citep{zadeh2017scalable},
\begin{small}
\begin{align}
    \vi(\bm{w}_i,\bm{w}_j)=& 1-\frac{ \ii(\bm{w}_i;\bm{w}_j) }{ \hh(\bm{w}_i,\bm{w}_j) } \nonumber\\
    =& 1- \frac{ \sum_{w\in\bm{w}_i, y\in\mathbf{y}}p(w,y)\log\frac{p(w,y)}{p(w)p(y)} }{ -\sum_{w\in\bm{w}_i, y\in\mathbf{y}}p(w,y)\log p(w,y) } \,,
\end{align}%
\end{small}%
is used to indicate the diversity between them.
The objective function for handling the trade-off between diversity and accuracy of two sub-architectures is defined as
\begin{equation}
\small
    \mathcal{L}_p(\mathbf{w}_i,\mathbf{w}_j)=
    (1-\alpha)\vi(\bm{w}_i,\bm{w}_j) + \alpha\frac{ \mi(\bm{w}_i,\mathbf{y}) + \mi(\bm{w_j},\mathbf{y}) }{2}
    \,, \label{eq:10}
\end{equation}
if $\mathbf{w}_i\cdot\mathbf{h}_i \neq \mathbf{w}_j\cdot\mathbf{h}_j$, otherwise $\mathcal{L}_p(\mathbf{w}_i,\mathbf{w}_j)=0\,.$
Note that $\alpha$ is a regularization factor introduced to balance between these two criteria, indicating their importance as well.
Our target is to pick up the $\mathbf{w}$ and $\mathbf{h}$, and prune them by minimizing 
$\mathcal{L}_e(\mathbf{w})$ in Eq.~(\ref{eq:11}), \ie{}
\begin{equation}
\small
    \mathcal{L}_e(\mathbf{w}_i)=
    \sum_{\mathbf{w}_j\cdot\mathbf{h}_j \in f\setminus\{\mathbf{w}_i\cdot\mathbf{h}_i\}}
    \mathcal{L}_p(\mathbf{w}_i,\mathbf{w}_j)
    \,. \label{eq:11}
\end{equation}
This loss function considers both diversity and accuracy concurrently according to the essential characteristics in ensemble learning.

\section{Experimental Study}
\label{experts}

In this section, we describe the experiments to verify the
effectiveness of the proposed \propabbr{} method. There are four
major questions that we aim to answer. (1) Could \propabbr{} achieve
sub-ensemble architectures at a smaller size yet still with
comparable accuracy performance to the original ensemble
architecture? (2) Could \propabbr{} generate sub-ensemble
architectures with more diversity than the original ensemble
architecture? (3) What are the impacts of the parameter $\alpha$ on
the sub-ensemble architectures generated by \infoabbr{}? (4) Could
\infoabbr{} generate different sub-architectures from that in the
original ensemble architecture?

\begin{table*}[!t]
\centering
\caption{%
Empirical results of ensemble-architectures' performance for binary classification on CIFAR-10, Fashion-MNIST, and MNIST datasets. 
Each method includes three columns, \ie{} the \emph{test accuracy (\%)},
the \emph{size} of generated (sub-)ensemble architectures, 
and the \emph{time cost (min)} of the searching process. 
The best of them are indicated with bold fonts for each label pair (row).
Note that sub-architectures used in these experiments are MLPs. 
}\label{tab:renew,binary}
\vspace{-1.5em}
\subfloat[Comparison on the \emph{test accuracy (\%)} performance.]{%
\scalebox{0.569}{
\begin{threeparttable}[b]
\begin{tabular}{l|cccccccc}%
\toprule
\multirow{2}{*}{\bf Label Pair} & \multicolumn{8}{c}{\bf Test Accuracy (\%)} \\
~ & \adanet{}&\randabbr{}&\valuabbr{}&\infoabbr{} & \adanvari{}&\randvari{}&\valuvari{}&\infovari{} \\
\midrule
digits 6-9       &  99.85$\pm$0.07 & 99.83$\pm$0.05 & 99.85$\pm$0.07 & 99.85$\pm$0.09 & 99.84$\pm$0.05 & \textbf{99.88$\pm$0.04}$\ddagger$ & 99.84$\pm$0.14$\dagger$ & 99.83$\pm$0.07 \\
digits 5-8       &  99.14$\pm$0.13 & 99.18$\pm$0.21 & 99.19$\pm$0.18 & 99.21$\pm$0.10$\ddagger$ & 99.16$\pm$0.18 & \textbf{99.26$\pm$0.15} & 99.19$\pm$0.18 & 99.20$\pm$0.15 \\
top-pullover     &  97.03$\pm$0.47 & 97.06$\pm$0.15$\ddagger$ & 97.03$\pm$0.39$\ddagger$ & 97.04$\pm$0.34$\ddagger$ & 97.04$\pm$0.30$\ddagger$ & \textbf{97.16$\pm$0.18}$\ddagger$ & 97.08$\pm$0.21$\ddagger$ & 96.94$\pm$0.17 \\
top-coat         &  98.62$\pm$0.14 & 98.57$\pm$0.29$\dagger$ & 98.61$\pm$0.07 & 98.64$\pm$0.32 & 98.61$\pm$0.22$\dagger$ & 98.61$\pm$0.22$\dagger$ & 98.66$\pm$0.23 & \textbf{98.67$\pm$0.14} \\
top-shirt        &  85.87$\pm$0.77 & 86.36$\pm$0.82 & 86.33$\pm$0.77$\ddagger$ & 86.18$\pm$0.80 & 86.14$\pm$0.54$\ddagger$ & \textbf{86.48$\pm$0.62}$\ddagger$ & 86.27$\pm$0.69$\ddagger$ & 86.36$\pm$0.54$\ddagger$ \\
trouser-dress    &  98.35$\pm$0.16 & \textbf{98.42$\pm$0.16} & 98.39$\pm$0.17 & 98.38$\pm$0.28 & 98.30$\pm$0.14 & 98.39$\pm$0.14$\ddagger$ & 98.41$\pm$0.27 & 98.32$\pm$0.24$\dagger$ \\
sandal-ankle boot&  98.74$\pm$0.19 & 98.78$\pm$0.27 & \textbf{98.79$\pm$0.24} & 98.71$\pm$0.11 & 98.78$\pm$0.14$\ddagger$ & 98.71$\pm$0.10 & 98.69$\pm$0.19 & 98.69$\pm$0.31$\dagger$ \\
deer-truck       &  87.91$\pm$0.38 & 88.01$\pm$0.42 & 87.87$\pm$0.85$\dagger$ & 87.99$\pm$0.67 & 87.95$\pm$0.35$\ddagger$ & \textbf{88.05$\pm$0.47} & 87.93$\pm$0.40 & 87.91$\pm$0.49 \\
deer-horse       &  75.54$\pm$1.45 & 76.22$\pm$1.17$\ddagger$ & 76.22$\pm$1.07$\ddagger$ & \textbf{76.62$\pm$0.94}$\ddagger$ & 76.10$\pm$1.28$\ddagger$ & 76.31$\pm$1.15$\ddagger$ & 76.22$\pm$0.41$\ddagger$ & 76.25$\pm$0.46$\ddagger$ \\
automobile-truck &  72.93$\pm$0.24 & 72.82$\pm$0.57$\dagger$ & 72.91$\pm$0.29$\dagger$ & 72.84$\pm$0.85$\dagger$ & 72.58$\pm$0.94$\dagger$ & 72.78$\pm$0.50$\dagger$ & 72.90$\pm$0.81$\dagger$ & \textbf{72.95$\pm$1.10} \\
cat-dog          &  61.15$\pm$0.69 & 61.11$\pm$0.23 & 61.02$\pm$1.24$\dagger$ & 60.67$\pm$1.05$\dagger$ & \textbf{61.63$\pm$0.81} & 61.60$\pm$0.68$\ddagger$ & 61.22$\pm$0.47$\ddagger$ & 61.53$\pm$1.34 \\
dog-horse        &  78.21$\pm$0.30 & 77.95$\pm$1.02$\dagger$ & 78.29$\pm$0.61 & \textbf{78.44$\pm$0.23}$\ddagger$ & 78.20$\pm$0.81$\dagger$ & 78.41$\pm$0.71 & 78.23$\pm$0.98 & 78.38$\pm$0.95 \\
\midrule
$t$-test (W/T/L) &  --- & 3/7/2 & 3/6/3 & 2/6/4 & 3/4/5 & 2/4/6 & 2/6/4 & 2/8/2   \\
Average Rank     &  5.71 & 4.75 & 4.71 & 3.92 & 5.58 & 2.67 & 4.25 & 4.42      \\
\bottomrule
\end{tabular}
\begin{tablenotes}
\item[1] The reported results are the average values of each method and the corresponding standard deviation under 5-fold cross-validation on each dataset. 
\item[2] By two-tailed paired $t$-test at 5\% significance level, $\ddagger$ and $\dagger$ denote that the performance of \adanet{} is inferior to and superior to that of the comparative \propabbr{} method with their variants, respectively. 
\item[3] The last two rows show the results of $t$-test and average rank, respectively. 
The ``W/T/L'' in $t$-test indicates that \adanet{} is superior to, not significantly different from, or inferior to the corresponding comparative \propabbr{} methods including their variants. The average rank is calculated according to the Friedman test~\cite{demvsar2006statistical}. 
\end{tablenotes}
\end{threeparttable}
}} \\
\vspace{-.2em}
\subfloat[Comparison on the \emph{size} of the (sub-)ensemble architectures.]{%
\scalebox{0.569}{
\begin{tabular}{l|cccccccc}
\toprule
\multirow{2}{*}{\bf Label Pair} & \multicolumn{8}{c}{\bf Number of Sub-Architectures} \\
~ & \adanet{}&\randabbr{}&\valuabbr{}&\infoabbr{} & \adanvari{}&\randvari{}&\valuvari{}&\infovari{} \\
\midrule
digits 6-9       &  \textbf{4.20$\pm$1.47} & 5.80$\pm$0.40 & 6.00$\pm$0.89 & 5.80$\pm$0.98 & 6.20$\pm$0.98 & 5.20$\pm$1.17 & 5.60$\pm$0.49 & 5.00$\pm$1.79$\dagger$ \\
digits 5-8       &  6.80$\pm$0.40 & 6.00$\pm$0.63$\ddagger$ & 6.60$\pm$0.49 & 5.80$\pm$0.75 & 6.00$\pm$0.63$\ddagger$ & 6.20$\pm$1.17 & \textbf{5.60$\pm$0.49}$\ddagger$ & 6.00$\pm$1.10 \\
top-pullover     &  5.00$\pm$0.63 & 5.40$\pm$0.49 & 5.80$\pm$0.98$\dagger$ & 5.20$\pm$0.98$\dagger$ & 5.00$\pm$0.89 & 3.80$\pm$1.17 & 4.20$\pm$0.75 & \textbf{3.20$\pm$0.40}$\ddagger$ \\
top-coat         &  5.40$\pm$0.80 & 4.80$\pm$0.40$\ddagger$ & 4.60$\pm$0.80$\ddagger$ & 5.40$\pm$0.80 & 5.20$\pm$0.75$\ddagger$ & 5.40$\pm$0.49$\ddagger$ & 4.40$\pm$0.80 & \textbf{3.00$\pm$0.00}$\ddagger$ \\
top-shirt        &  5.60$\pm$0.49 & 5.40$\pm$0.80 & 5.60$\pm$0.80 & 5.20$\pm$1.47 & 5.60$\pm$1.02 & 5.80$\pm$0.75$\dagger$ & \textbf{4.20$\pm$0.98} & 4.60$\pm$1.62 \\
trouser-dress    &  4.20$\pm$1.47 & 5.20$\pm$0.75 & 5.20$\pm$1.17 & 4.40$\pm$1.36 & 5.00$\pm$1.10 & 4.00$\pm$1.79 & \textbf{4.00$\pm$0.63}$\ddagger$ & 4.60$\pm$1.62$\dagger$ \\
sandal-ankle boot&  5.20$\pm$0.75 & 5.80$\pm$1.17$\dagger$ & 5.40$\pm$1.02$\dagger$ & 5.40$\pm$1.36$\dagger$ & 6.20$\pm$0.75$\dagger$ & 5.40$\pm$0.49 & 4.80$\pm$0.75$\ddagger$ & \textbf{3.40$\pm$0.80}$\ddagger$ \\
deer-truck       &  4.80$\pm$1.17 & 4.80$\pm$1.17 & 5.00$\pm$0.89 & 4.60$\pm$1.02$\ddagger$ & 5.20$\pm$0.75 & 5.20$\pm$1.33$\dagger$ & 4.60$\pm$0.80$\ddagger$ & \textbf{4.20$\pm$0.98}$\ddagger$ \\
deer-horse       &  4.00$\pm$0.63 & 4.40$\pm$0.80$\dagger$ & 5.20$\pm$1.17$\dagger$ & \textbf{3.40$\pm$0.80} & 5.00$\pm$0.00$\dagger$ & 5.00$\pm$0.00$\dagger$ & 5.00$\pm$0.63$\dagger$ & 5.20$\pm$0.75$\dagger$ \\
automobile-truck &  4.40$\pm$1.02 & 4.20$\pm$1.47 & 4.40$\pm$0.80$\ddagger$ & \textbf{3.20$\pm$1.33} & 5.00$\pm$1.41$\dagger$ & 5.00$\pm$1.26$\dagger$ & 5.20$\pm$0.40 & 4.80$\pm$0.75 \\
cat-dog          &  4.00$\pm$1.10 & 4.00$\pm$1.26 & 4.00$\pm$1.26 & 4.40$\pm$1.50$\dagger$ & \textbf{3.40$\pm$0.49}$\ddagger$ & 5.40$\pm$0.80 & 4.60$\pm$1.02 & 3.60$\pm$0.49$\ddagger$ \\
dog-horse        &  \textbf{4.00$\pm$1.10} & 5.00$\pm$0.89 & 5.40$\pm$1.02 & 5.00$\pm$0.63 & 4.60$\pm$0.80 & 5.00$\pm$0.63$\dagger$ & 4.20$\pm$0.75 & 5.40$\pm$0.80 \\
\midrule
$t$-test (W/T/L) &  --- & 2/8/2 & 3/7/2 & 3/8/1 & 3/6/3 & 5/6/1 & 1/7/4 & 3/4/5 \\
Average Rank     &  3.96 & 4.79 & 6.00 & 4.00 & 5.38 & 5.38 & 3.25 & 3.25 \\
\bottomrule
\end{tabular}
}} \\
\vspace{-.7em}
\subfloat[Comparison on the \emph{time cost (min)} of the searching process.]{%
\scalebox{0.569}{
\begin{tabular}{l|cccccccc}
\toprule
\multirow{2}{*}{\bf Label Pair} & \multicolumn{8}{c}{\bf Time Cost (min)} \\
~ & \adanet{}&\randabbr{}&\valuabbr{}&\infoabbr{} & \adanvari{}&\randvari{}&\valuvari{}&\infovari{} \\
\midrule
digits 6-9       &  \textbf{10.92$\pm$1.82} & 12.35$\pm$0.99  & 12.91$\pm$1.31  & 13.04$\pm$0.76 & 12.74$\pm$1.29 & 12.24$\pm$0.92 & 13.50$\pm$0.28 & 12.61$\pm$1.84$\dagger$ \\
digits 5-8       &  13.00$\pm$0.46 & 12.25$\pm$0.64$\ddagger$  & 13.96$\pm$0.34$\dagger$  & \textbf{12.16$\pm$0.49}$\ddagger$ & 13.07$\pm$0.46$\dagger$ & 13.07$\pm$0.83$\dagger$ & 13.16$\pm$0.62$\dagger$ & 13.55$\pm$1.18$\dagger$ \\
top-pullover     &  11.83$\pm$0.46 & 11.93$\pm$0.52$\dagger$  & 12.78$\pm$1.29$\dagger$  & 11.73$\pm$1.77 & 11.84$\pm$0.98$\dagger$ & 11.22$\pm$1.01 & 12.08$\pm$0.58$\dagger$ & \textbf{11.02$\pm$0.28}$\ddagger$ \\
top-coat         &  12.00$\pm$0.68 & 11.21$\pm$0.34$\ddagger$  & \textbf{11.05$\pm$0.70}  & 11.78$\pm$1.27 & 12.42$\pm$0.52 & 11.90$\pm$0.67$\ddagger$ & 12.04$\pm$0.62 & 11.09$\pm$0.17$\ddagger$ \\
top-shirt        &  11.80$\pm$1.06 & 12.06$\pm$1.18$\dagger$  & 12.63$\pm$0.85  & \textbf{10.89$\pm$2.53} & 12.55$\pm$0.66 & 12.63$\pm$0.74 & 12.04$\pm$0.83 & 12.26$\pm$1.34$\dagger$ \\
trouser-dress    &  10.88$\pm$1.40 & 11.98$\pm$0.61  & 12.36$\pm$1.02$\dagger$  & 11.24$\pm$1.74$\dagger$ & 12.19$\pm$0.66 & \textbf{10.75$\pm$1.72} & 11.92$\pm$0.71 & 12.25$\pm$1.23 \\
sandal-ankle boot&  11.75$\pm$0.94 & 12.09$\pm$0.99$\dagger$  & 12.55$\pm$1.04$\dagger$  & \textbf{10.20$\pm$2.35} & 13.03$\pm$0.40 & 12.34$\pm$0.47 & 12.20$\pm$0.77 & 11.34$\pm$0.91$\ddagger$ \\
deer-truck       &  15.50$\pm$1.81 & 13.32$\pm$2.08  & 11.72$\pm$1.01$\ddagger$  & \textbf{11.39$\pm$0.81}$\ddagger$ & 16.74$\pm$1.00 & 16.11$\pm$1.24 & 16.90$\pm$0.90 & 15.51$\pm$1.18 \\
deer-horse       &  14.66$\pm$1.09 & 12.70$\pm$0.98$\ddagger$  & 11.94$\pm$1.08$\ddagger$  &  \textbf{9.82$\pm$0.58}$\ddagger$ & 16.22$\pm$0.68 & 15.99$\pm$1.08 & 16.94$\pm$1.04 & 17.12$\pm$1.03$\dagger$ \\
automobile-truck &  15.31$\pm$1.33 & 12.64$\pm$2.23  & 11.35$\pm$1.15$\ddagger$  & \textbf{10.04$\pm$1.31}$\ddagger$ & 15.87$\pm$1.70$\dagger$ & 16.42$\pm$1.36$\dagger$ & 16.52$\pm$0.50 & 16.97$\pm$0.89 \\
cat-dog          &  24.28$\pm$17.37 & 17.17$\pm$1.48$\ddagger$ & 23.06$\pm$11.87$\ddagger$ & 75.08$\pm$113.67$\dagger$ & \textbf{14.45$\pm$1.29}$\ddagger$ & 78.41$\pm$104.25$\dagger$ & 113.55$\pm$193.13$\dagger$ & 35.34$\pm$38.05$\dagger$ \\
dog-horse        &  16.79$\pm$2.68 & 23.08$\pm$12.98$\dagger$ & 77.02$\pm$116.93$\dagger$ & \textbf{16.07$\pm$1.02}$\ddagger$ & 71.94$\pm$108.17$\dagger$ & 119.54$\pm$189.82$\dagger$ & 17.00$\pm$1.81 & 46.69$\pm$39.39$\dagger$ \\
\midrule
$t$-test (W/T/L) &  --- & 4/4/4 & 5/3/4 & 2/5/5 & 4/7/1 & 4/7/1 & 3/9/0 & 6/3/3 \\
Average Rank     &  3.25 & 3.50 & 5.17 & 2.50 & 5.58 & 5.00 & 6.08 & 4.92 \\
\bottomrule
\end{tabular}
}}
\end{table*}

\subsection{Three Image Classification Datasets}

The three image classification datasets that we employ in the
experiments are all publicly available.
The ImageNet \citep{deng2009imagenet} dataset is not included since the cost for it is not affordable for one GPU (NVIDIA GTX 1080) that we use.

\noindent\textbf{CIFAR-10} \citep{krizhevsky2009learning}: %
60,000 32x32 color images in 10 classes are used as instances, with 6,000 images per class, representing airplanes, automobiles, birds, cats, deer, dogs, frogs, horses, ships, and trucks, respectively. There are 50,000 training images and 10,000 test images.

\noindent\textbf{MNIST} \citep{lecun1998gradient}: %
70,000 28x28 grayscale images of handwritten digits in 10 different classes are used as instances. There are 60,000 instances as a training set and 10,000 instances as a test set. The digits have been size-normalized and centered in a fixed-size image.

\noindent\textbf{Fashion-MNIST} \citep{xiao2017fashion}: %
70,000 28x28 grayscale images are used as instances, including 60,000 instances for training and 10,000 instances for testing. They are categorized into ten classes, representing T-shirts/tops, trousers, pullovers, dresses, coats, sandals, shirts, sneakers, bags, and ankle boots, respectively.

\begin{figure}
\centering
\subfloat[]{\centering\label{fig,new:dnn,accuracy,a}
    \includegraphics[scale=0.56]{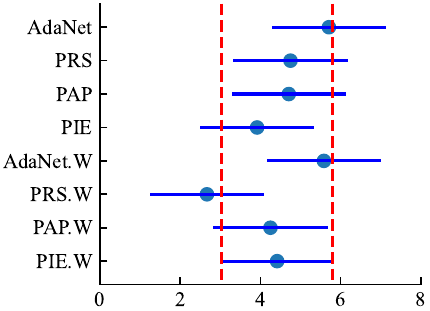}}
\subfloat[]{\centering\label{fig,new:dnn,accuracy,b}
    \includegraphics[scale=0.56]{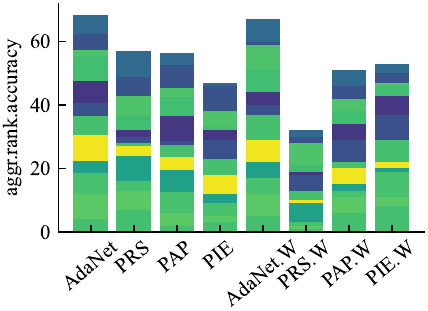}}
\caption{%
Comparison of the baseline \adanet{} and the proposed \propabbr{} including their variants on the \emph{test accuracy}, using MLPs as sub-architectures for binary classification. 
(a) Friedman test chart (non-overlapping means significant difference) \cite{demvsar2006statistical}, which rejects the assumption that ``all methods have the same accuracy performance'' at the significance level of $10\%$.  
(b) The aggregated rank of test accuracy for each method (the smaller the better) \cite{qian2015pareto}.
}\label{fig,new:dnn,accuracy}
\end{figure}

\subsection{Baseline Methods}
To analyze the effectiveness of \propabbr{}, we compare the three proposed solutions  (\ie{} \randabbr{}, \valuabbr{}, and \infoabbr{}) with \adanet{} \citep{cortes2017adanet}.
Besides, \adanet{} (usually set to use uniform average weights in practice) has a variant to use mixture weights, which we call \adanvari{} \citep{weill2018adanet}. 
Similarly, \randvari{}, \valuvari{}, and \infovari{} (\ie{} \propvari{}) are variants of \randabbr{}, \valuabbr{}, and \infoabbr{} using mixture weights, respectively.
Our baselines include \adanet{} and their corresponding variants. 
Besides, to objectively evaluate the performance of these methods, standard $5$-fold cross-validation is used in these experiments, \ie{} in each iteration, the entire data set is split into two parts, with $80\%$ as the training set and $20\%$ as the test set.

\begin{table}[tbhp]
\centering
\caption{%
Empirical results of ensemble-architectures' performance for multi-class classification. 
Each method includes four columns, \ie{} the \emph{test accuracy (\%)},
the \emph{size} of generated (sub-)ensemble architectures, 
the \emph{diversity} of the pruned sub-ensemble architectures, 
and the \emph{time cost (min)} of the searching process. 
The best of them are indicated with bold fonts for each dataset (row).
Note that sub-architectures used in these experiments are MLPs and CNNs. 
}\label{tab:refresh,multi}
\renewcommand\tabcolsep{1.7pt}%
\scalebox{0.54}{%
\begin{threeparttable}[b]
\begin{tabular}{l|cccccccc}
\toprule
\multirow{2}{*}{\bf Dataset} & \multicolumn{8}{c}{\bf Test Accuracy (\%)} \\
~ & \adanet{}&\randabbr{}&\valuabbr{}&\infoabbr{} & \adanvari{}&\randvari{}&\valuvari{}&\infovari{} \\
\midrule
MNIST            &  94.88$\pm$0.22  & 94.82$\pm$0.36$\dagger$ & 94.79$\pm$0.23$\dagger$ & 94.75$\pm$0.25$\dagger$  & 94.94$\pm$0.29  & 94.66$\pm$0.13   & 94.56$\pm$0.28$\dagger$   & \textbf{94.94$\pm$0.19}$\ddagger$  \\
Fashion-MNIST    &  83.74$\pm$0.52  & 83.76$\pm$0.88 & \textbf{83.95$\pm$0.50}$\ddagger$ & 83.89$\pm$0.64  & 83.81$\pm$0.32$\ddagger$  & 83.98$\pm$0.40$\ddagger$   & \textbf{84.24$\pm$0.18}$\ddagger$   & 83.93$\pm$0.21$\ddagger$  \\
MNIST$^*$        &  90.54$\pm$0.24 & 90.46$\pm$0.25$\dagger$ & 90.44$\pm$0.15 & 90.35$\pm$0.24$\dagger$ & \textbf{90.55$\pm$0.18}$\ddagger$ & 90.38$\pm$0.27$\dagger$ & 90.27$\pm$0.16 & 90.23$\pm$0.35$\dagger$  \\
Fashion-MNIST$^*$& 81.39$\pm$0.43 & \textbf{81.48$\pm$0.30}$\ddagger$ & 81.40$\pm$0.23$\ddagger$ & 81.32$\pm$0.45$\dagger$ & 81.39$\pm$0.26$\ddagger$ & 81.41$\pm$0.18$\ddagger$ & 81.20$\pm$0.09 & 81.05$\pm$0.58$\dagger$  \\
\midrule
$t$-test (W/T/L) &  --- & 2/1/1 & 1/1/2 & 3/1/0 & 0/1/3 & 1/1/2 & 1/2/1 & 2/0/2 \\
Average Rank     &  4.50 & 3.75 & 3.75 & 5.75 & 3.25 & 4.00 & 5.75 & 5.25 \\
\bottomrule
\toprule
\multirow{2}{*}{\bf Dataset} & \multicolumn{8}{c}{\bf Number of Sub-Architectures} \\
~ & \adanet{}&\randabbr{}&\valuabbr{}&\infoabbr{} & \adanvari{}&\randvari{}&\valuvari{}&\infovari{}  \\
\midrule
MNIST            &  6.80$\pm$0.40 & 6.60$\pm$0.49 & 7.00$\pm$0.00 & 6.60$\pm$0.49 & 6.60$\pm$0.49 & 6.20$\pm$0.98 & \textbf{5.20$\pm$0.75}$\ddagger$ & 6.80$\pm$0.40  \\
Fashion-MNIST    &  5.40$\pm$1.02 & 5.60$\pm$0.80 & 5.40$\pm$1.02 & 6.00$\pm$0.63 & 6.00$\pm$0.89 & 5.60$\pm$0.80 & 5.00$\pm$0.63$\ddagger$ & \textbf{4.00$\pm$1.10}  \\
MNIST$^*$        &  5.80$\pm$0.75  & 4.80$\pm$0.75$\ddagger$  & 5.60$\pm$1.50  & 4.80$\pm$0.40$\ddagger$  & 5.40$\pm$1.36  & 5.00$\pm$1.41  & 3.80$\pm$1.17  & \textbf{3.00$\pm$0.00}$\ddagger$  \\
Fashion-MNIST$^*$&  5.40$\pm$1.36  & 3.80$\pm$1.47$\ddagger$  & 6.40$\pm$0.49  & 5.00$\pm$0.63$\ddagger$  & 5.60$\pm$0.49  & 4.00$\pm$0.63$\ddagger$  & 4.00$\pm$0.89$\ddagger$  & \textbf{3.20$\pm$0.40}$\ddagger$  \\
\midrule
$t$-test (W/T/L) &  --- & 0/2/2	& 0/4/0 & 0/2/2 & 0/4/0 & 0/3/1 & 0/1/3 & 0/2/2 \\
Average Rank     &  6.00 & 3.75 & 6.63 & 5.00 & 6.13 & 4.00 & 2.13 & 2.38  \\
\bottomrule
\toprule
\multirow{2}{*}{\bf Dataset} & \multicolumn{8}{c}{\bf Time Cost (min)} \\
~ & \adanet{}&\randabbr{}&\valuabbr{}&\infoabbr{} & \adanvari{}&\randvari{}&\valuvari{}&\infovari{} \\
\midrule
MNIST            &  78.65$\pm$89.02 & 23.42$\pm$2.32$\ddagger$ & \textbf{21.51$\pm$0.16}$\ddagger$ & 31.73$\pm$12.46$\ddagger$ & 81.13$\pm$87.66 & 43.36$\pm$12.72$\ddagger$  & 92.39$\pm$117.81$\dagger$ & 33.11$\pm$1.94$\ddagger$ \\
Fashion-MNIST    &  21.75$\pm$1.87  & \textbf{20.07$\pm$0.65}$\ddagger$ & 25.50$\pm$9.07$\dagger$ & 39.30$\pm$6.03$\dagger$  & 42.92$\pm$9.71$\dagger$  & 91.41$\pm$117.32$\dagger$ & 31.77$\pm$2.16$\dagger$   & 75.92$\pm$89.19$\dagger$ \\
MNIST$^*$        &  30.73$\pm$1.10 & 29.03$\pm$0.74$\ddagger$ & 28.04$\pm$3.87 & \textbf{20.47$\pm$1.27}$\ddagger$ & 30.46$\pm$1.72 & 25.88$\pm$2.62$\ddagger$ & 29.08$\pm$1.23 & 28.25$\pm$0.08$\ddagger$  \\
Fashion-MNIST$^*$&  29.61$\pm$1.53 & 27.74$\pm$1.75$\ddagger$ & 30.63$\pm$1.53$\dagger$ & \textbf{20.90$\pm$0.83}$\ddagger$ & 31.12$\pm$0.57$\dagger$ & 25.01$\pm$3.19 & 28.22$\pm$1.92 & 28.28$\pm$0.19$\ddagger$  \\
\midrule
$t$-test (W/T/L) &  --- & 0/0/4 & 2/1/1 & 1/0/3 & 2/2/0 & 1/1/2 & 2/2/0 & 1/0/3 \\
Average Rank     &  5.50 & 2.75 & 3.50 & 2.50	& 7.00 & 4.25 & 5.50 & 5.00 \\
\bottomrule
\end{tabular}
\begin{tablenotes}
\item[1] Empirical results with $^*$ represent experiments using CNNs as sub-architectures;  
Empirical results without $^*$ represent experiments using MLPs as sub-architectures. 
\item[2] The reported results are the average values of each method and the corresponding standard deviation under 5-fold cross-validation on each dataset. 
\item[3] By two-tailed paired $t$-test at 5\% significance level, $\ddagger$ and $\dagger$ denote that the performance of \adanet{} is inferior to and superior to that of the comparative \propabbr{} method with their variants, respectively. 
\item[4] The last two rows show the results of $t$-test and average rank, respectively. 
The ``W/T/L'' in $t$-test indicates that \adanet{} is superior to, not significantly different from, or inferior to the corresponding comparative \propabbr{} methods including their variants. The average rank is calculated according to the Friedman test~\cite{demvsar2006statistical}. 
\end{tablenotes}
\end{threeparttable}
}
\end{table}

\subsection{Experimental Settings}

In the same experiment, all methods would use the same kind of sub-architectures in consideration of fairness during the comparisons to verify whether their objectives work well.
The optional sub-architectures that we use include multilayer perceptrons (MLPs) and convolutional neural networks (CNNs).
Note that those CNNs are only composed of convolution layers with 16 channels, without pooling layers.
As for the hyper-parameters in the experiments, the learning rate is set to be $0.003$, 
and cosine decay is applied to the learning rate using a momentum optimizer in the training process.
The number of training steps is $5,000$, and that of the batch size is $64$. 

We use three datasets mentioned before for image classification.
In the multi-class classification scenario, we use all of the categories in the corresponding dataset;
in the binary classification scenario, we reduce these datasets by considering several pairs of classes.
For example, we consider five pairs of classes in CIFAR-10 (\ie{} deer-truck, deer-horse, automobile-truck, cat-dog, and dog-horse), 
five pairs of classes in Fashion-MNIST (\ie{} top-pullover, top-coat, top-shirt, trouser-dress, and sandal-ankle boot), and two pairs of digits in MNIST (\ie{} digits $6$-$9$, and $5$-$8$).

\begin{figure}
\centering
\subfloat[][\centering\label{fig,new:dnn,arch,time,a}]{
\includegraphics[scale=0.56]{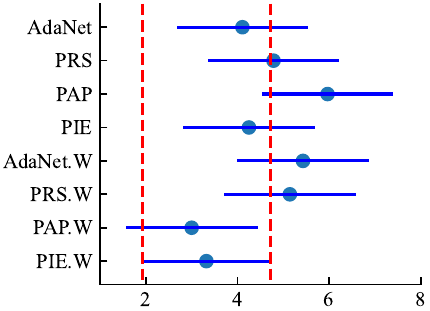}}
\subfloat[][\centering\label{fig,new:dnn,arch,time,b}]{
\includegraphics[scale=0.56]{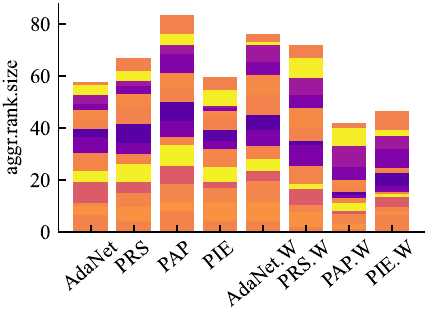}}
\\
\subfloat[][\centering\label{fig,new:dnn,arch,time,c}]{
\includegraphics[scale=0.56]{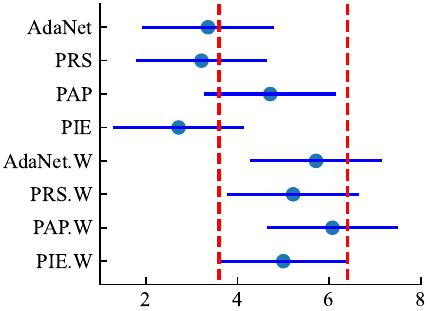}}
\subfloat[][\centering\label{fig,new:dnn,arch,time,d}]{
\includegraphics[scale=0.56]{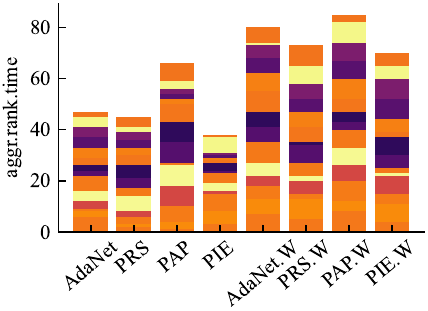}}
\caption{%
Comparison of the baseline \adanet{} and the proposed \propabbr{} including their variants, using MLPs as sub-architectures for image classification. 
(a--b) Comparison on the \emph{size} of generated (sub-)ensemble architectures. 
(c--d) Comparison on the \emph{time cost} of the searching process. 
Notice that: the Friedman test chart in (a) rejects the assumption that ``the size of ensemble architectures of different methods has no significant difference'' at $5\%$ significant level; 
that in (c) rejects the assumption that ``the time cost of different methods has no significant difference'' at $5\%$ significant level. 
}\label{fig,new:dnn,arch,time}
\end{figure}

\begin{figure*}
\centering %
\subfloat[][\centering\label{fig,new:acc,expts,a}]{
\includegraphics[scale=0.67]{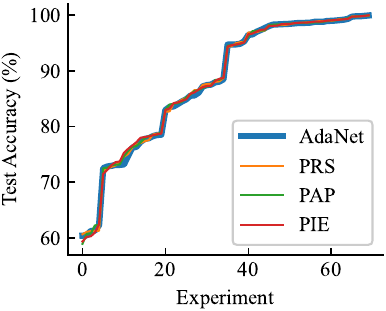}}
\hspace{4ex}
\subfloat[][\centering\label{fig,new:acc,expts,b}]{
\includegraphics[scale=0.67]{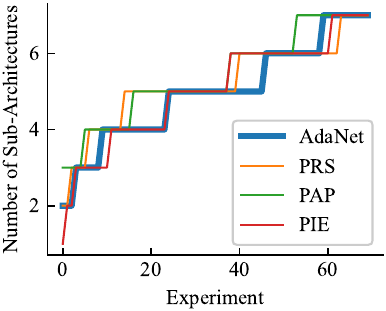}}
\hspace{4ex}
\subfloat[][\centering\label{fig,new:acc,expts,c}]{
\includegraphics[scale=0.67]{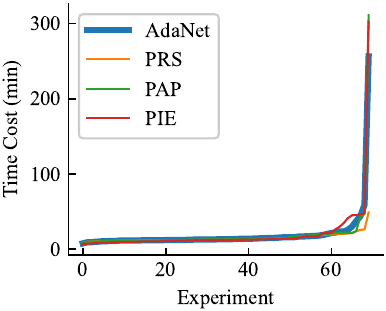}}
\\
\subfloat[][\centering\label{fig,new:acc,expts,d}]{
\includegraphics[scale=0.67]{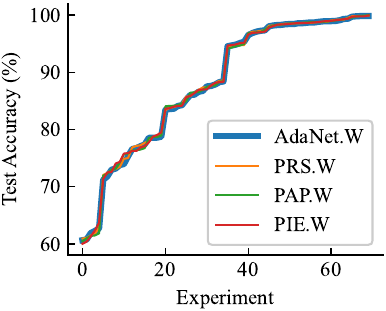}}
\hspace{4ex}
\subfloat[][\centering\label{fig,new:acc,expts,e}]{
\includegraphics[scale=0.67]{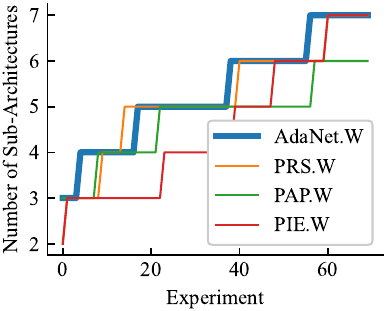}}
\hspace{4ex}
\subfloat[][\centering\label{fig,new:acc,expts,f}]{
\includegraphics[scale=0.67]{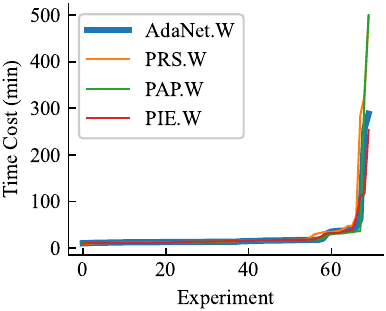}}
\caption{%
Comparison of the baseline \adanet{} and the proposed \propabbr{} including their corresponding variants, using MLPs as sub-architectures for image classification. 
The horizontal axis represents empirical results in different specific experiments like different rows in Tables~\ref{tab:renew,binary}--\ref{tab:refresh,multi}. 
(a--c) Comparison of performance of \adanet{} and \propabbr{}. 
(d--f) Comparison of performance of their corresponding variants.
}\label{fig,new:acc,expts}
\end{figure*}

\subsection{\propabbr{} Could Achieve Ensemble Architectures with Better Performance of Accuracy}%
In this subsection, we verify whether the pruned sub-ensemble architectures could achieve comparable performance with the original ensemble architecture.
Experimental results are reported in Tables~\ref{tab:renew,binary}--\ref{tab:refresh,multi} contain the average test accuracy (\%) of each method and the corresponding standard deviation under $5$-fold cross-validation on each data set. 
For instance, each row (data set) in Table~\ref{tab:renew,binary} compares the classification accuracy using sub-architectures with the same type, 
indicating results with higher accuracy and lower standard deviation by bold fonts. 
When comparing one method with \adanet{}, the one with higher values of accuracy and lower standard deviation would win; otherwise, the winner would be decided based on the significance of the difference in the accuracy performance between the two methods, which is examined by two-tailed paired $t$-test at $5\%$ significance level to tell if two methods have significantly different results. 
Specifically, 
two methods end up with a tie if there is no significant statistical difference between them; otherwise, the one with higher values of accuracy would win. 
The performance of each method is reported in the last two rows of Table~\ref{tab:renew,binary}, compared with \adanet{} in terms of the average rank and the number of data sets that \adanet{} has won, tied, or lost, respectively. 
We may notice that \propabbr{} achieved better results than \adanet{} in most cases, yet with possible larger time cost in a few cases.
Therefore, it could be referred that \propabbr{} could generate ensemble architectures with better performance of accuracy. 
Figure~\ref{fig,new:dnn,accuracy}\subref{fig,new:dnn,accuracy,a} shows that \propabbr{} (indicated by \randabbr{}, \valuabbr{}, and \infoabbr{}) achieves the same level of accuracy performance as \adanet{} at least, and their variants even exhibits better accuracy performance than \adanet{} and \adanvari{}. 
Similar results are presented in Figure~\ref{fig,new:dnn,accuracy}\subref{fig,new:dnn,accuracy,b} and Table~\ref{tab:refresh,multi}.

\subsection{\propabbr{} Leads to Ensemble Architectures with Smaller Size}
In this subsection, we verify whether the pruned sub-ensemble architectures could generate comparable performance of architectures with smaller size. 
Experimental results are reported in Tables~\ref{tab:renew,binary}--\ref{tab:refresh,multi} and Figures~\ref{fig,new:dnn,arch,time}--\ref{fig,new:acc,expts}. 
As we can see in Table~\ref{tab:renew,binary}, \propabbr{} achieves ensemble architectures with the
smallest size in most cases, although the significant difference between different methods might not be as large as that of accuracy, as shown in Figures~\ref{fig,new:dnn,arch,time}\subref{fig,new:dnn,arch,time,a}--\ref{fig,new:dnn,arch,time}\subref{fig,new:dnn,arch,time,b}. 
Meanwhile, Table~\ref{tab:renew,binary} and Figures~\ref{fig,new:dnn,arch,time}\subref{fig,new:dnn,arch,time,c}--\ref{fig,new:dnn,arch,time}\subref{fig,new:dnn,arch,time,d} presents that variants of \adanet{} and \propabbr{} might cost more time than themselves. 
However, considering that \propabbr{} already achieves the comparable performance with \adanet{} and that \propabbr{} generates ensemble architectures with smaller size indeed, we believe that our NAS ensemble pruning method is still meaningful somehow. 
Similar observations are exhibited in Figure~\ref{fig,new:acc,expts} as well: 
(1) \propabbr{} could achieve the same level of accuracy performance as \adanet{}, as shown in Figures~\ref{fig,new:acc,expts}\subref{fig,new:acc,expts,a} and \ref{fig,new:acc,expts}\subref{fig,new:acc,expts,d}; 
(2) \propabbr{} could generate ensemble architectures with competitive performance yet smaller size, as shown in Figure~\ref{fig,new:acc,expts}\subref{fig,new:acc,expts,e}.

\begin{table*}[tbhp]
\centering
\caption{%
Empirical results under different $\alpha$ values on the MNIST dataset for binary classification (to be specific, the label pair of digits $5-8$), using MLPs as sub-architectures. 
Each method includes four columns: the \emph{test accuracy (\%)}, the \emph{diversity (disagreement)}, the \emph{size} (\ie{} the number of sub-architectures), and the \emph{time cost (min)} of the searching process. 
Note that ``orig.'' represents \adanet{}, \randabbr{}, \valuabbr{}, or \infoabbr{}; ``vari.'' represents \adanvari{}, \randvari{}, \valuvari{}, or \infovari{}, correspondingly. 
}\label{tab,renew:alpha}
\begin{threeparttable}[b]
\scalebox{.67}{
\begin{tabular}{l|cc|cc|cc|cc}
\toprule
\multirow{2}{*}{\bf } & \multicolumn{2}{c|}{\bf Test Accuracy (\%)} & \multicolumn{2}{c|}{\bf Diversity (Disagreement)} & \multicolumn{2}{c|}{\bf Size} & \multicolumn{2}{c}{\bf Time Cost (min)} \\
~ & orig. & vari. & orig. & vari. & orig. & vari. & orig. & vari. \\
\midrule
\midrule
\adanet{}  &  99.86$\pm$0.06 & 99.86$\pm$0.05 & 0.0003$\pm$0.0001 & 0.0005$\pm$0.0003 & 5.60$\pm$5.60 & 5.80$\pm$5.80 & 11.53$\pm$0.43 & 12.37$\pm$0.75 \\
\randabbr{}&  99.83$\pm$0.07 & 99.88$\pm$0.05 & 0.0037$\pm$0.0043 & 0.0019$\pm$0.0031 & 5.40$\pm$5.40 & 4.80$\pm$4.80 & 11.93$\pm$0.71 & 11.16$\pm$1.52 \\
\valuabbr{}&  99.87$\pm$0.05 & 99.88$\pm$0.04 & 0.0011$\pm$0.0018 & 0.0003$\pm$0.0001 & 5.80$\pm$5.80 & 5.40$\pm$5.40 & 12.76$\pm$1.06 & 12.89$\pm$0.62 \\
\infoabbr{} ($\alpha=0.5$) %
           &  99.80$\pm$0.06 & 99.85$\pm$0.04 & 0.0037$\pm$0.0069 & 0.0030$\pm$0.0053 & 4.80$\pm$4.80 & 5.80$\pm$5.80 & 12.29$\pm$1.14 & 13.15$\pm$0.54 \\
\midrule
\infoabbr{} ($\alpha=0.0$) &  99.15$\pm$0.23 & 99.19$\pm$0.26 & 0.0022$\pm$0.0004 & 0.0399$\pm$0.0201 & 7.00$\pm$0.00 & 5.80$\pm$0.40 & 14.53$\pm$0.10 & 13.98$\pm$0.63 \\
\infoabbr{} ($\alpha=0.05$)&  99.16$\pm$0.24 & 99.16$\pm$0.13 & 0.0024$\pm$0.0004 & 0.0280$\pm$0.0135 & 6.60$\pm$0.80 & 6.00$\pm$1.10 &  9.56$\pm$2.14 & 13.60$\pm$1.16 \\
\infoabbr{} ($\alpha=0.1$) &  99.13$\pm$0.04 & 99.21$\pm$0.15 & 0.0019$\pm$0.0003 & 0.0456$\pm$0.0250 & 6.40$\pm$0.49 & 5.40$\pm$1.20 & 13.59$\pm$0.80 & 13.22$\pm$0.91 \\
\infoabbr{} ($\alpha=0.15$)&  99.24$\pm$0.14 & 99.25$\pm$0.17 & 0.0020$\pm$0.0004 & 0.0313$\pm$0.0299 & 6.00$\pm$0.89 & 5.80$\pm$0.98 & 12.95$\pm$0.64 & 14.00$\pm$0.75 \\
\infoabbr{} ($\alpha=0.2$) &  99.14$\pm$0.17 & 99.21$\pm$0.06 & 0.0022$\pm$0.0006 & 0.0574$\pm$0.0162 & 6.20$\pm$0.75 & 6.40$\pm$0.49 & 14.20$\pm$0.72 & 14.58$\pm$0.18 \\
\infoabbr{} ($\alpha=0.25$)&  99.22$\pm$0.12 & 99.15$\pm$0.19 & 0.0020$\pm$0.0004 & 0.0378$\pm$0.0294 & 6.00$\pm$0.63 & 6.40$\pm$0.80 & 13.08$\pm$0.61 & 14.30$\pm$0.71 \\
\infoabbr{} ($\alpha=0.3$) &  99.29$\pm$0.12 & 99.18$\pm$0.22 & 0.0023$\pm$0.0003 & 0.0415$\pm$0.0251 & 7.00$\pm$0.00 & 6.40$\pm$0.49 & 13.16$\pm$1.54 & 14.36$\pm$0.41 \\
\infoabbr{} ($\alpha=0.35$)&  99.19$\pm$0.11 & 99.24$\pm$0.09 & 0.0021$\pm$0.0006 & 0.0210$\pm$0.0158 & 5.60$\pm$1.02 & 5.40$\pm$1.20 & 12.85$\pm$0.73 & 13.20$\pm$0.93 \\
\infoabbr{} ($\alpha=0.4$) &  99.22$\pm$0.13 & 99.20$\pm$0.22 & 0.0022$\pm$0.0002 & 0.0364$\pm$0.0288 & 6.80$\pm$0.40 & 6.20$\pm$0.75 & 11.02$\pm$0.12 & 14.49$\pm$0.53 \\
\infoabbr{} ($\alpha=0.45$)&  99.12$\pm$0.12 & 99.18$\pm$0.18 & 0.0023$\pm$0.0009 & 0.0486$\pm$0.0232 & 6.60$\pm$0.80 & 6.60$\pm$0.80 & 13.62$\pm$0.40 & 14.59$\pm$0.44 \\
\infoabbr{} ($\alpha=0.55$)&  99.17$\pm$0.19 & 99.18$\pm$0.20 & 0.0022$\pm$0.0004 & 0.0454$\pm$0.0107 & 6.00$\pm$0.63 & 6.40$\pm$0.49 & 12.99$\pm$0.69 & 14.45$\pm$0.59 \\
\infoabbr{} ($\alpha=0.6$) &  99.23$\pm$0.08 & 99.19$\pm$0.15 & 0.0022$\pm$0.0002 & 0.0499$\pm$0.0275 & 6.00$\pm$1.10 & 5.60$\pm$0.80 &  9.96$\pm$1.46 & 13.65$\pm$1.03 \\
\infoabbr{} ($\alpha=0.65$)&  99.20$\pm$0.14 & 99.29$\pm$0.12 & 0.0018$\pm$0.0003 & 0.0173$\pm$0.0147 & 6.40$\pm$0.80 & 5.40$\pm$1.02 & 13.14$\pm$0.87 & 13.21$\pm$1.08 \\
\infoabbr{} ($\alpha=0.7$) &  99.23$\pm$0.21 & 99.25$\pm$0.12 & 0.0020$\pm$0.0003 & 0.0234$\pm$0.0228 & 6.60$\pm$0.80 & 6.00$\pm$1.26 &  8.17$\pm$0.64 & 13.78$\pm$0.83 \\
\infoabbr{} ($\alpha=0.75$)&  99.16$\pm$0.21 & 99.22$\pm$0.15 & 0.0022$\pm$0.0005 & 0.0428$\pm$0.0216 & 6.60$\pm$0.49 & 6.00$\pm$0.63 & 13.48$\pm$0.51 & 14.21$\pm$0.64 \\
\infoabbr{} ($\alpha=0.8$) &  99.17$\pm$0.15 & 99.19$\pm$0.03 & 0.0028$\pm$0.0017 & 0.0269$\pm$0.0213 & 6.40$\pm$0.49 & 5.40$\pm$1.20 &  7.80$\pm$0.47 & 13.35$\pm$1.05 \\
\infoabbr{} ($\alpha=0.85$)&  99.20$\pm$0.13 & 99.18$\pm$0.17 & 0.0020$\pm$0.0003 & 0.0411$\pm$0.0224 & 5.60$\pm$0.49 & 6.60$\pm$0.49 & 12.81$\pm$0.35 & 14.45$\pm$0.47 \\
\infoabbr{} ($\alpha=0.9$) &  99.20$\pm$0.23 & 99.28$\pm$0.16 & 0.0017$\pm$0.0003 & 0.0596$\pm$0.0172 & 6.00$\pm$0.63 & 6.00$\pm$0.63 &  7.67$\pm$0.92 & 13.99$\pm$0.45 \\
\infoabbr{} ($\alpha=0.95$)&  99.22$\pm$0.19 & 99.22$\pm$0.10 & 0.0020$\pm$0.0003 & 0.0283$\pm$0.0210 & 5.80$\pm$0.40 & 5.80$\pm$1.17 & 12.51$\pm$0.38 & 13.82$\pm$0.80 \\
\infoabbr{} ($\alpha=1.0$) &  99.19$\pm$0.19 & 99.28$\pm$0.15 & 0.0021$\pm$0.0006 & 0.0483$\pm$0.0159 & 6.20$\pm$0.75 & 6.40$\pm$0.49 &  8.07$\pm$0.35 & 14.34$\pm$0.27 \\
\bottomrule
\end{tabular}
}
\end{threeparttable}
\end{table*}

\begin{figure*}
\centering %
\subfloat[][\centering\label{fig,new3:acc,expts,a}]{
\includegraphics[scale=0.57]{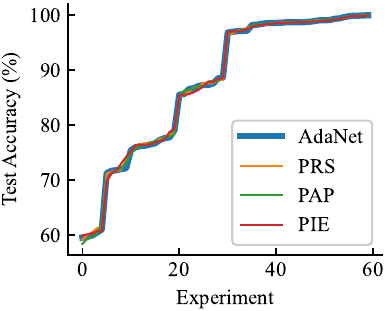}}
\hspace{2ex}
\subfloat[][\centering\label{fig,new3:acc,expts,g}]{
\includegraphics[scale=0.57]{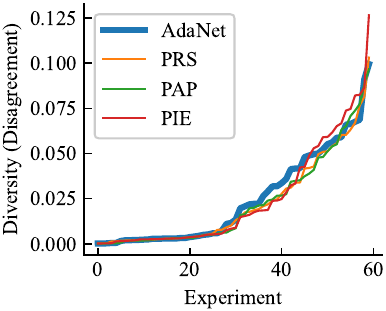}}
\hspace{2ex}
\subfloat[][\centering\label{fig,new3:acc,expts,b}]{
\includegraphics[scale=0.57]{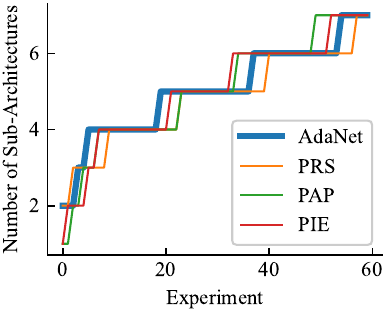}}
\hspace{2ex}
\subfloat[][\centering\label{fig,new3:acc,expts,c}]{
\includegraphics[scale=0.57]{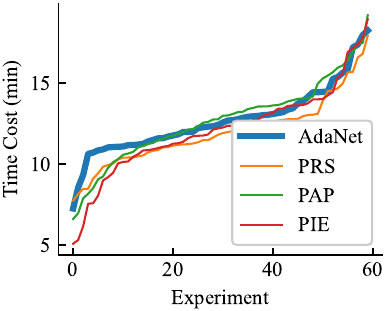}}
\\
\subfloat[][\centering\label{fig,new3:acc,expts,d}]{
\includegraphics[scale=0.57]{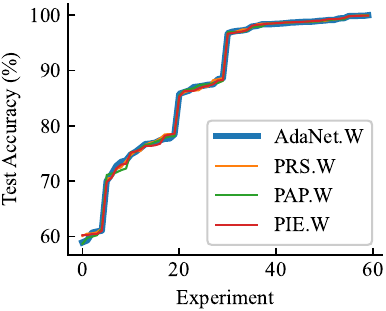}}
\hspace{2ex}
\subfloat[][\centering\label{fig,new3:acc,expts,h}]{
\includegraphics[scale=0.57]{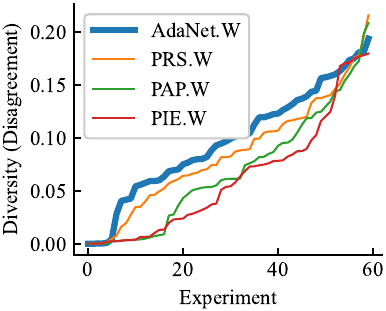}}
\hspace{2ex}
\subfloat[][\centering\label{fig,new3:acc,expts,e}]{
\includegraphics[scale=0.57]{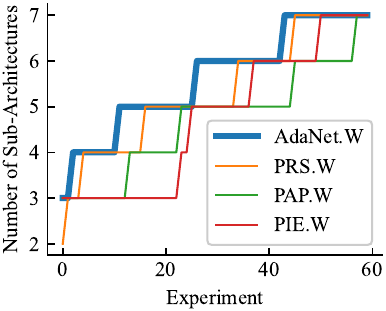}}
\hspace{2ex}
\subfloat[][\centering\label{fig,new3:acc,expts,f}]{
\includegraphics[scale=0.57]{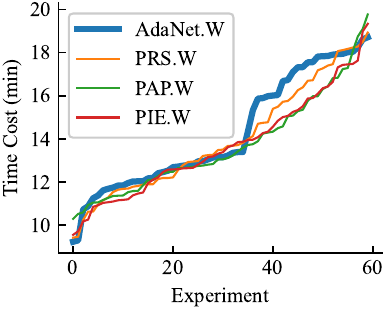}}
\caption{%
Comparison of the baseline \adanet{} and the proposed \propabbr{} including their corresponding variants, using MLPs as sub-architectures for binary classification. 
The horizontal axis represents empirical results in different specific experiments like different rows in Tables~\ref{tab:renew,binary}--\ref{tab:refresh,multi}. 
(a--c) Comparison of performance of \adanet{} and \propabbr{}. 
(d--f) Comparison of performance of their corresponding variants.
}\label{fig,new3:acc,expts}
\end{figure*}

\begin{figure*}
\centering
\subfloat[][\centering\label{fig,new:alpha,effect,a}]{
\includegraphics[scale=0.44]{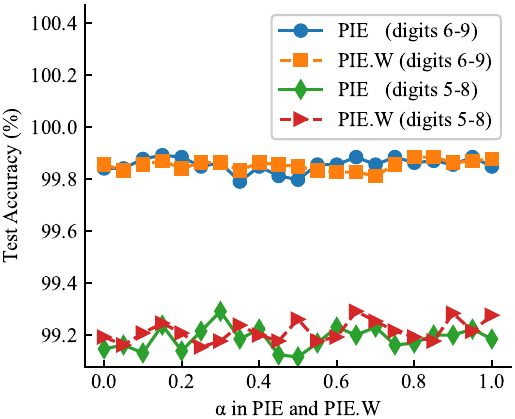}}
\hspace{2ex}
\subfloat[][\centering\label{fig,new:alpha,effect,b}]{
\includegraphics[scale=0.44]{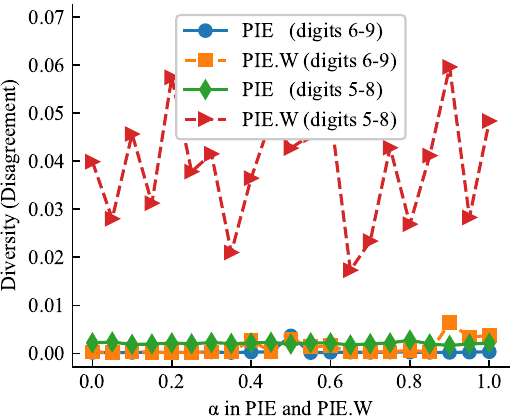}}
\hspace{2ex}
\hspace{2ex}
\subfloat[][\centering\label{fig,new:alpha,effect,c}]{
\includegraphics[scale=0.44]{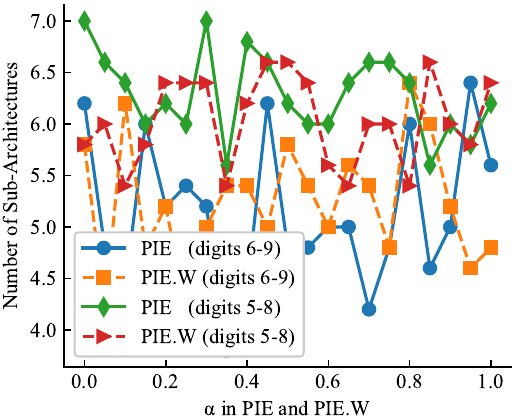}}
\hspace{2ex}
\subfloat[][\centering\label{fig,new:alpha,effect,d}]{
\includegraphics[scale=0.44]{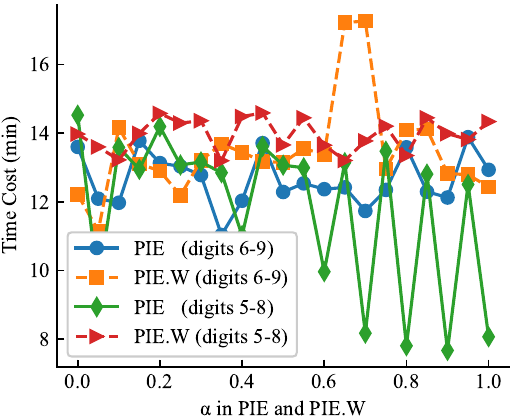}}
\caption{%
The effect of different $\alpha$ values in \infoabbr{} and \infovari{} for binary classification. 
(a) The effect of the $\alpha$ value on the test accuracy performance of sub-ensemble architectures. 
(b) The effect of the $\alpha$ value on the diversity of sub-ensemble architectures, measured by the disagreement measure in Eq.~\eqref{eq:16}. 
(c) The effect of the $\alpha$ value on the size of sub-ensemble architectures. 
(d) The effect of the $\alpha$ value on the time cost.
}\label{fig,new:alpha,effect}
\end{figure*}

\subsection{\infoabbr{} Generates Sub-Ensemble Architectures with More Diversity}
\label{eg:diver}

In this subsection, we verify whether the purpose of increasing the diversity of ensemble architectures is satisfied.
We use the normalized of information $\vi$ in \infoabbr{} to imply the redundancy between two different sub-architectures, indicating the diversity between them.
However, in this experiment, we use another measure named the disagreement measure \citep{skalak1996sources,ho1998random} here to calculate the diversity for the ensemble architecture and the pruned sub-ensemble architectures, because there is no analogous term like $\vi$ in \randabbr{} and \valuabbr{}.
Note that researchers proposed many other measures to calculate diversity, and the disagreement measure is one of them \citep{zhou2012ensemble}.
We choose the disagreement measure here because this measure is easy to be calculated and understood.
The disagreement between two sub-architectures $\mathbf{w}_i$ and $\mathbf{w}_j$ is
\begin{equation}
\small
    \disagreement(\mathbf{w}_i,\mathbf{w}_j)=
    \frac{1}{m}\sum_{1\leqslant i\leqslant m} \mathbb{I}(\mathbf{h}_i(\bm{x}_i) \neq \mathbf{h}_j(\bm{x}_i))
    \,,\label{eq:15}
\end{equation}
the diversity of the ensemble architecture $f$ using the disagreement measure is 
\begin{equation}
\small
    \bm{\disagreement}(f)=\frac{2}{l(l-1)} \sum_{\mathbf{w}_i\cdot\mathbf{h}_i\in f} \sum_{\substack{\mathbf{w}_j\cdot\mathbf{h}_j\in f,\\ \mathbf{h}_j\neq\mathbf{h}_i}} \disagreement(\mathbf{w}_i,\mathbf{w}_j)
    \,,\label{eq:16}
\end{equation}
and the diversity of the sub-ensemble architecture $f\setminus\{\mathbf{w}\cdot\mathbf{h}\}$ could be calculated analogously.

Table~\ref{tab,renew:alpha} and Figures~\ref{fig,new3:acc,expts}--\ref{fig,new:alpha,effect} 
report their performance with the corresponding disagreement value reflecting the diversity of the whole ensemble architecture as well.
Besides, 
Table~\ref{tab,renew:alpha} reports the diversity of the
sub-architectures using \infoabbr{} and other corresponding
information. Note that the larger the disagreement is, the larger
the diversity of the ensemble architecture or the pruned sub-architecture is. 
\valuabbr{} in Table~\ref{tab,renew:alpha} achieves better accuracy performance and more diversity concurrently. 
Similar results are observed in \randvari{} and \valuvari{} compared with \adanvari{} in Table~\ref{tab,renew:alpha}, which illustrates that the accuracy of the sub-ensemble architecture could benefit from increasing diversity. 
Meanwhile, Table~\ref{tab,renew:alpha} shows that larger sub-ensemble architectures correspond to less diversity sometimes. 
In addition, Figure~\ref{fig,new:alpha,effect} indicate the effect of the $\alpha$ value in Eq.~(\ref{eq:10}) on the diversity, the accuracy performance, the time cost, and the size of the sub-ensemble architectures.

\subsection{Effect of the $\alpha$ Value}
\label{eg:alpha}

This subsection will investigate the effect of the hyper-parameter
$\alpha$ in \infoabbr{}. The value of $\alpha$ indicates the
relation between two criteria in Eq.~(\ref{eq:10}) as well. To
reveal this issue, different $\alpha$ values (from 0.0 to 1.0 with
0.05 steps) are evaluated in the experiments. 
Figure~\ref{fig,new:alpha,effect} exemplify the effect of $\alpha$ on the MNIST dataset, taking the label pairs of digits $5$-$8$ and digits $6$-$9$ as an example. 
Figure~\ref{fig,new:alpha,effect}\subref{fig,new:alpha,effect,a} illustrates that the accuracy of sub-ensemble architectures is affected slightly under different $\alpha$ values yet would not cause much accuracy decline. 
Figure~\ref{fig,new:alpha,effect}\subref{fig,new:alpha,effect,b} illustrates that the diversity of the sub-ensemble architectures in \infovari{} is affected under different $\alpha$ values yet without large changes of absolute values; meanwhile, the diversity of that in \infoabbr{} is almost not affected under different $\alpha$ values. 
Figures~\ref{fig,new:alpha,effect}\subref{fig,new:alpha,effect,c}--\ref{fig,new:alpha,effect}\subref{fig,new:alpha,effect,d} present that the size and time cost of sub-ensemble architectures would be more affected under different $\alpha$ values. 
Generally, the size and time cost of sub-ensemble architectures in \infoabbr{} tend to be decreased with the increase of $\alpha$ value. 

\begin{table}[t!]
\small\centering
\caption{
Empirical results of the exact sub-architectures that are kept in the final sub-ensemble architecture after pruning on the digits $5$-$8$ label pair in the MNIST dataset. 
Each method includes five columns: the \emph{test accuracy (\%)}, the \emph{diversity (disagreement)}, the \emph{time cost (min)}, the \emph{size} (\ie{} the number of sub-architectures), and the \emph{indexes} of the generated sub-architectures. 
Note that the sub-architectures used in these experiments are MLPs.
}\label{tab:distinct}
\scalebox{0.81}{%
\begin{tabular}{l|ccccl}
\toprule
~ & Accuracy & Diversity & Time & Size & Indexes  \\
\midrule
\adanet{}                  & 99.86 & 0.0001 & 10.71 & 6 & [0,1,3,4,5,6] \\
\randabbr{}                & 99.92 & 0.0101 & 10.54 & 4 & [0,1,2,5] \\
\valuabbr{}                & 99.93 & 0.0002 & 13.95 & 7 & [0,1,2,3,4,5,6] \\
\infoabbr{} ($\alpha$=0.25)& 99.24 & 0.0016 & 13.31 & 6 & [0,1,2,3,4,5] \\
\infoabbr{} ($\alpha$=0.5) & 99.89 & 0.0002 & 12.89 & 5 & [0,1,2,3,4] \\
\infoabbr{} ($\alpha$=0.75)& 99.12 & 0.0024 & 13.53 & 6 & [0,1,2,3,5,6] \\
\adanvari{}                & 99.82 & 0.0002 & 11.69 & 5 & [0,1,2,3,6] \\
\randvari{}                & 99.93 & 0.0006 & 13.82 & 7 & [0,1,2,3,4,5,6] \\
\valuvari{}                & 99.86 & 0.0003 & 12.47 & 6 & [0,1,3,4,5,6] \\
\infovari{} ($\alpha$=0.25)& 99.20 & 0.0029 & 12.91 & 5 & [0,1,2,3,6] \\
\infovari{} ($\alpha$=0.5) & 99.78 & 0.0004 & 13.69 & 6 & [0,1,2,3,4,5] \\
\infovari{} ($\alpha$=0.75)& 99.01 & 0.0445 & 13.99 & 6 & [0,1,2,3,4,6] \\
\bottomrule
\end{tabular}
}
\end{table}

\subsection{\adanet{} v.s. \propabbr{} over the Time Cost}
In this subsection, we compare the time cost of \adanet{} and \propabbr{} with their corresponding variants.
Experimental results are reported in 
Tables~\ref{tab:renew,binary}--\ref{tab,renew:alpha} and Figures~\ref{fig,new:dnn,arch,time}--\ref{fig,new:alpha,effect}, containing the accuracy on the test set of each method and their corresponding time cost. 
Although Figures~\ref{fig,new:dnn,arch,time}\subref{fig,new:dnn,arch,time,c}--\ref{fig,new:dnn,arch,time}\subref{fig,new:dnn,arch,time,d} illustrate that the time cost is not an advantage of \propabbr{} compared with \adanet{} while achieving the same level of accuracy, 
Tables~\ref{tab:renew,binary}--\ref{tab:refresh,multi} present that 
\propabbr{} could generate satisfactory sub-ensemble architectures within less time sometimes.
Generally, the time cost depends on the number of sub-architectures that are generated during the entire searching process, although the pruning is proceeded through the same process. 
Therefore, it is quite understandable that \propabbr{} might take a longer time if more sub-architectures are generated during searching. 
Moreover, Figure~\ref{fig,new:alpha,effect}\subref{fig,new:alpha,effect,d} presents the effect of different $\alpha$ values in \infoabbr{} on the time cost of generating sub-ensemble architectures with more diversity.

\subsection{\propabbr{} Could Generate Distinct Deeper Sub-Architectures than \adanet{}}
In a few cases, we observe that \infoabbr{} could achieve a larger ensemble architecture than \adanet{}, which makes us wonder whether \propabbr{} could lead to distinct architectures from \adanet{}.
Thus, we dig the sub-architectures that are kept in the final architecture to explore more details deep down inside. 
As we can see in Table~\ref{tab:distinct}, the size of sub-ensemble architectures tends to be larger under the lower level of diversity. 
The reason why \infoabbr{} (or \infovari{}) generates distinct deeper sub-architectures might be the diversity is not sufficient for its objective in Eq.~(\ref{eq:11}).
In this case, the objective would guide the pruning process to search for more distinct deeper sub-architectures to increase diversity.

\section{Related Work}
\label{related}

In this section, we introduce the neural architecture search (NAS) briefly.
The concept of ``neural architecture search (NAS)'' was proposed by Zoph~and~Le~\citep{zoph2017neural} for the very first time. 
They presented NAS as a gradient-based method to find good
architectures. A ``controller'', denoted by a recurrent network, was
used to generate variable-length string which specified the
structure and connectivity of a neural network; the generated
``child network,'' specified by the string, was then trained on the
real data to obtain accuracy as the reward signal, to generate an
architecture with higher probabilities to receive high accuracy
\citep{zoph2017neural,baker2016designing,zoph2018learning}. Existing
NAS methods could be categorized under three dimensions: search
space, search strategy, and performance estimation strategy
\citep{elsken2019neural,kandasamy2018neural,cai2018efficient,liu2018darts}.
Classical NAS methods yielded chain-structured neural architectures
\citep{zela2018towards,elsken2019neural}, yet ignored some modern
designed elements from hand-crafted architectures, such as skip
connections from ResNet \citep{he2016deep}. Thus some researchers
also attempted to build complex multi-branch networks by
incorporating those and achieved positive results
\citep{cai2018path,real2018regularized,elsken2018efficient,brock2017smash,elsken2017simple,zhong2018practical,pham2018efficient,zhong2018blockqnn}.

Recently, NAS methods involved ensemble learning are attracting researchers' attention gradually.
Cortes~\etal{}~\citep{cortes2017adanet} 
proposed a data-dependent learning guarantee to guide the choice of additional sub-networks and presented \adanet{} to learn neural networks adaptively.
They claimed that \adanet{} could precisely address some of the issues of wasteful data, time, and resources in neural architecture search since their optimization problem for \adanet{} was convex and admitted a unique global solution.
Besides, Huang~\etal{}~\citep{huang2018learning} 
specialized sub-architectures by residual blocks and claimed that their BoostResNet boosted over multi-channel representations/features, which was different from AdaNet.
Macko~\etal{}~\citep{macko2019improving} 
also proposed another attempt named as AdaNAS to utilize ensemble methods to compose a neural network automatically, which was an extension of \adanet{} with the difference of using subnetworks comprising stacked NASNet \citep{zoph2017neural,zoph2018learning} blocks.
However, both of them gathered all searched sub-architectures together and missed out on the critical characteristic that ensemble models usually benefit from diverse individual learners.

Moreover, Chang~\etal{}~\citep{chang2019differentiable} proposed Differentiable ARchiTecture Search with Ensemble Gumbel-Softmax (DARTS-EGS) and developed ensemble Gumbel-Softmax to maintain efficiency in searching. 
Ardywibowo~\etal{}~\citep{ardywibowo2020nads} constructed an ensemble model to perform the Out-of-Distribution (OoD) detection in their Neural Architecture Distribution Search (NADS), which searched for a distribution of architectures instead of one single best-performing architecture in standard neural architecture search methods.
These two methods are not discussed in this paper since they are not assembling sub-architectures during searching.

\section{Conclusion}
\label{result}

Recent attempts on NAS with ensemble learning methods have achieved
prominent results in reducing the search complexity and improving
the effectiveness~\citep{cortes2017adanet}. However, current
approaches usually miss out on an essential characteristic of
diversity in ensemble learning. To tackle this problem, in this
paper, we target the ensemble learning methods in NAS and propose an
ensemble pruning method named ``\emph{\propfull{}
(}\propabbr{}\emph{)}'' to reduce the redundant sub-architectures
during the searching process. Three solutions are proposed as the
guiding criteria in \propabbr{} that reflect the characteristics of
the ensemble architecture (\ie{} \randabbr{}, \valuabbr{}, and
\infoabbr{}) to prune the less valuable sub-architectures.
Experimental results indicate that \propabbr{} could guide diverse
sub-architectures to create sub-ensemble architectures in a smaller
size yet still with comparable performance to the ensemble
architecture that is not pruned. Besides, \infoabbr{} might lead to
distinct deeper sub-architectures if diversity is insufficient. In
the future, we plan to generalize the current method to more diverse
ensemble strategies and derive theoretical guarantees to further
improve the performance of the NAS ensemble architectures.


%



\ifCLASSOPTIONcaptionsoff
  \newpage
\fi


\bibliographystyle{IEEEtran}
\bibliography{title_abbr,refs}
%

\end{document}